\documentclass[11pt,a4paper]{article}

\usepackage{xpeng-template}

\usepackage{hyperref}

\usepackage[percent]{overpic}
\graphicspath{{figures/}}

\definecolor{TableNote}{RGB}{70,70,70}
\NewDocumentCommand{\doug}
{ mO{} }{\textcolor{brown}{\textsuperscript{\textit{doug}}\textsf{\textbf{\small[#1]}}}}

\newcommand{\xvista}{Capek 0.5\xspace}

\newcommand{\xvistatwo}{Capek 0.5-2B\xspace}
\newcommand{\xvistalarge}{Capek 0.5-35B-A3B\xspace}
\newcommand{\qwentwob}{Qwen3.5-2B\xspace}
\newcommand{\qwenlarge}{Qwen3.6-35B-A3B\xspace}

\newcommand{\vigil}{VIGIL\xspace}

\xpengcorrespondence{Jie Chen}{chenj81@xiaopeng.com}

\title{{Capek 0.5}: An Execution-Centric Vision-Language Model \\ for Embodied Intelligence}

\author{%
{\bfseries
Ying Chen\textsuperscript{*},
Weizhen Li\textsuperscript{*},
Zhe Hu\textsuperscript{*},
Zhenjiang Li\textsuperscript{*},
Rui Jiang\textsuperscript{*},
Zhifeng Gu,
Lihuang Fang,
Jiangping Liu,
Lei Yi\textsuperscript{\textdagger},
Jie Chen\textsuperscript{\textdagger}}\\[0.1em]
{\small\textsuperscript{*}Core contributors.\quad
\textsuperscript{\textdagger}Correspondence.}%
}

\date{}  

\hypersetup{
  pdftitle={Capek 0.5: An Execution-Centric Vision-Language Model for Robot-Scene Reasoning},
  pdfauthor={Ying Chen, Weizhen Li, Zhe Hu, Zhenjiang Li, Rui Jiang, Zhifeng Gu, Lihuang Fang, Jiangping Liu, Lei Yi, Jie Chen},
  pdfsubject={Technical report on execution-centric vision-language modeling for robot-scene reasoning},
  pdfkeywords={vision-language model, robotics, robot-scene reasoning, spatial reasoning, temporal understanding, visual grounding, trajectory prediction, physical state}
}

\begin{document}

\begin{xpenghero}
  \begin{abstract}

Vision-language models are increasingly serving as the reasoning core of embodied agents. Robot execution is inherently iterative: each action reshapes the scene and physical state, continually renewing what must be perceived, reasoned about, and verified. Meeting these demands requires complementary capabilities that differ in supervision signals, prediction formats, and verification criteria. Existing approaches typically develop
these capabilities against isolated, task-specific objectives, leaving open how they should be organized and integrated around execution as a whole. We present \xvista, an embodied vision-language  model built around an execution-centric capability taxonomy. Rather than organizing training by datasets or tasks, the taxonomy groups embodied capabilities according to their functional roles throughout execution and comprises four capability families: Spatial Reasoning, Temporal Understanding, Action Guidance, and State Verification. Each capability is first acquired by a dedicated specialist through reinforcement learning with verifiable rewards from a shared backbone, and the specialists are then consolidated into a single inference-time model through weight-space merging followed by routed policy-space distillation.
We instantiate \xvista at the 2B and 35B-A3B scales and evaluate it from three complementary perspectives: comprehensive benchmark suites including Capek-StateBench, a new benchmark for state verification; a controlled study of capability retention from specialists to the unified model; and closed-loop evaluation in simulated embodied environments. \xvista improves the large majority of matched benchmark rows over its initialization, retains all four specialized capabilities in one checkpoint with quantified losses, and transfers to closed-loop embodied task execution. The results indicate that execution-centric specialization followed by consolidation is a practical recipe for building unified embodied vision-language models.

\end{abstract}

\end{xpenghero}

\vspace{-0.20em}
\begin{center}
\includegraphics[width=0.98\linewidth]{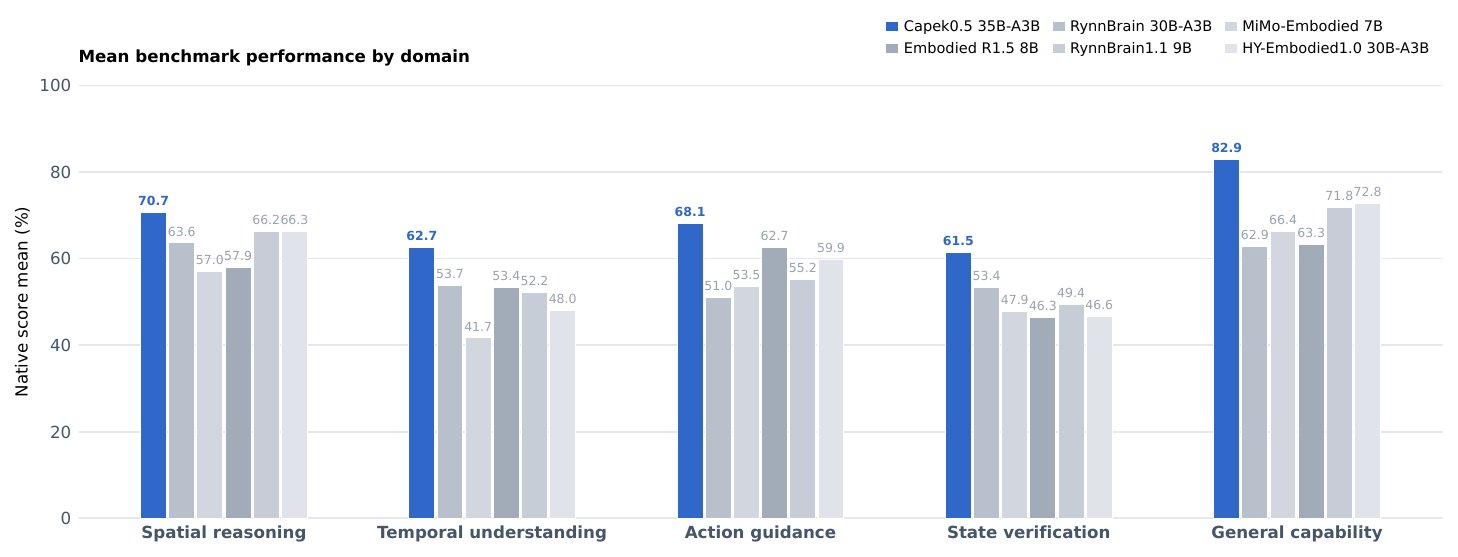}
\vspace{-0.5em}
\captionsetup{hypcap=false,font=footnotesize}
\captionof{figure}{\textbf{Average performance across capability domain.} Results from Table~\ref{tab:public_results_large} are averaged within five domains: Spatial Reasoning, Temporal Understanding, Action Guidance, State Verification, and General Capability. NaviTrace, VABench-trace, and ShareRobot-Trajectory are excluded from the Action Guidance average due to incompatible trajectory metrics. The results show that \xvistalarge maintains strong and balanced performance across embodied and general capabilities.}
\label{fig:benchmark_snapshot}
\end{center}
\vspace{-0.50em}

\newpage
{
  \setlength{\baselineskip}{0.8\baselineskip}
  \tableofcontents
}
\newpage

\section{Introduction}
\label{sec:intro}

Vision-language pretraining supplies strong perceptual and linguistic priors~\citep{radford2021clip,alayrac2022flamingo,li2023blip2,dai2023instructblip,liu2023llava,wang2024qwen2vl,bai2025qwen25vl}, enabling vision-language models (VLMs) to support semantic planning, visual reasoning, and language-based interfaces to downstream robot control~\citep{saycan2022,palme2023,codeaspolicies2022,voxposer2023,geminirobotics2025}. Recent robot-centric models further extend this foundation with complementary abilities such as affordance grounding, trajectory prediction,  temporal understanding, and physically grounded planning~\citep{robopoint2024,robobrain2025,robobrain25_2026,rynnbrain2026,hyembodied10_2026,cosmos3_2026,mimoembodied2025,embodiedr15,vesta2026,acebrain2026}, positioning VLMs as an ``embodied brain'' that perceives, reasons, and directs action for increasingly capable embodied agents~\citep{yang2025embodiedbench,salimpour2025towards}. 

Yet current models still face challenges because robot execution is inherently iterative: each action changes the scene and physical state, continually renewing the evidence required for subsequent decisions. Throughout this \textit{execution cycle}, a VLM must repeatedly reason about spatial context, interpret events over time, ground actions to the environment, and determine whether physical predicates hold and whether task goals have been achieved. 
Effective robot execution therefore relies on intermediate predictions that remain interpretable and verifiable, allowing each decision to reliably inform the next.
Existing post-training pipelines are commonly organized around individual tasks, datasets, or benchmark-specific objectives. Even when such supervision is mixed within a single model, heterogeneous output structures and scoring rules make it difficult to determine which execution capabilities have been acquired and whether they are retained after integration.

\begin{figure}[H]
\vspace{-3mm}
  \centering
  \includegraphics[width=\linewidth]{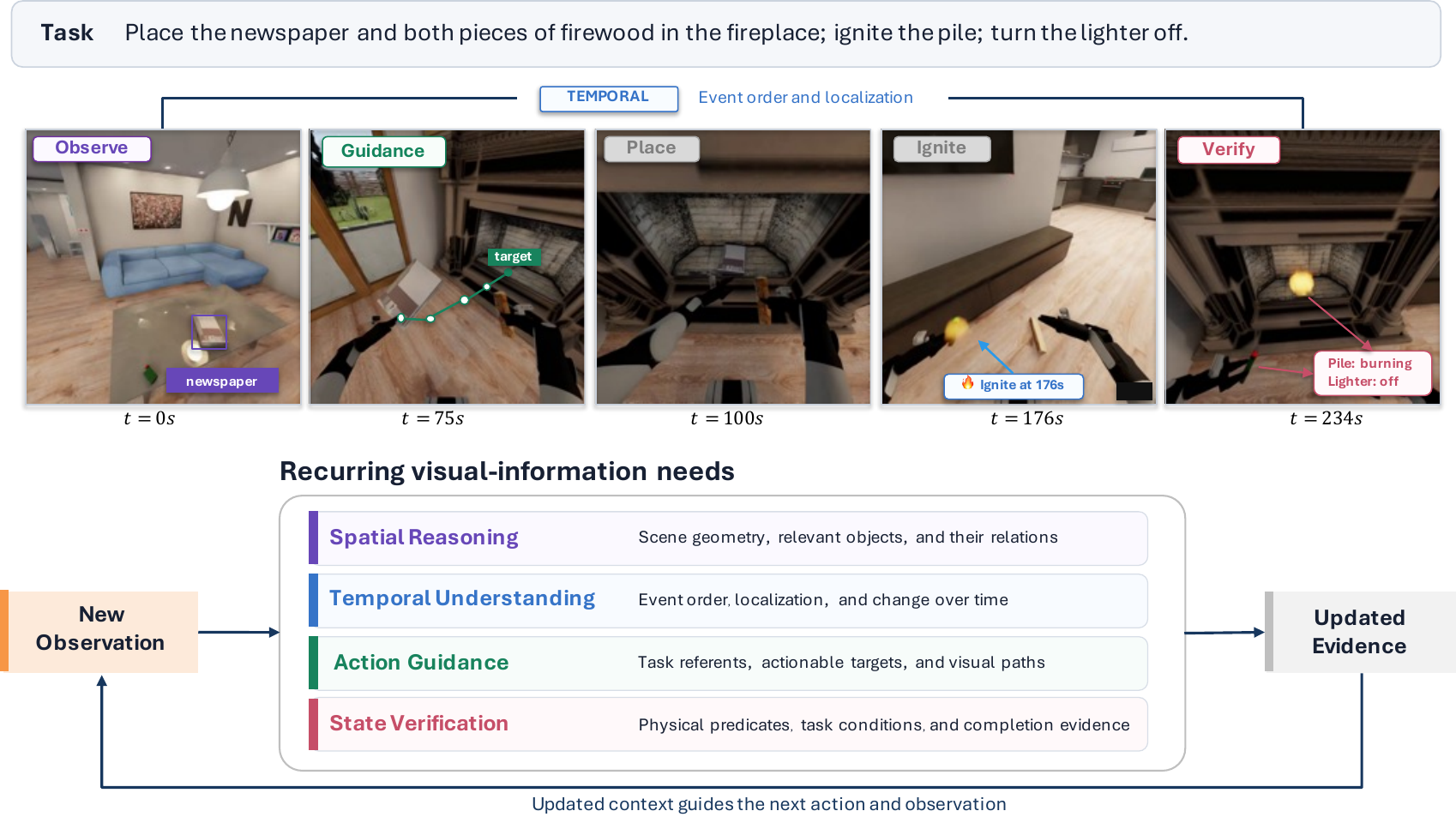}
  \vspace{-3mm}
  \caption{\textbf{Execution-centric capability view.} A BEHAVIOR-1K household manipulation trace~\citep{li2023behavior1k} illustrates how Spatial, Temporal, Guidance, and State requirements recur and overlap across an observation--reasoning--action cycle rather than form mutually exclusive stages.}
  \label{fig:execution_capability_lens}
  \vspace{-3mm}
\end{figure}

In this work, we present \xvista, a unified embodied VLM for robot perception and reasoning. Rather than organizing post-training solely around individual tasks or datasets, \xvista adopts an \emph{execution-centric} capability taxonomy that groups recurring capabilities according to their functional roles throughout robot execution. This taxonomy provides a common framework for training and evaluating capabilities that are repeatedly invoked as the robot perceives, acts, and verifies progress in dynamic environments.
Concretely, it comprises four capability families: \textbf{Spatial Reasoning} captures scene geometry and inter-entity relations; \textbf{Temporal Understanding} models event progression and identifies when relevant changes occur; \textbf{Action Guidance} grounds task-relevant entities and affordances and produces image-space points, regions, and ordered trajectories; and \textbf{State Verification} determines whether visually supported physical predicates hold and whether task goals have been achieved. 
Figure~\ref{fig:execution_capability_lens} illustrates how these four capabilities recur throughout a household manipulation episode as execution proceeds. 
 
Unifying this taxonomy within a single model, however, is nontrivial: the four capabilities differ sharply in their supervision signals and output formats, spanning structured relations, temporal localization, geometric grounding, and visually grounded predicates.
Such heterogeneous objectives make joint post-training with a single undifferentiated recipe susceptible to optimization interference across capabilities~\citep{mopd2026,wang2026mix}. 
To address this challenge, we separate capability acquisition from capability integration. All four specialists start from the same backbone checkpoint with capability-aligned data and objectives. Specialist checkpoints are then combined with TIES~\citep{ties2023} to initialize a unified student, which is further consolidated through routed multi-teacher on-policy distillation (MOPD)~\citep{mopd2026} on student-generated prefixes. 

We instantiate two model scales, \xvistalarge and \xvistatwo, from \qwenlarge\citep{qwen36_35b_a3b} and \qwentwob\citep{qwen3.5} respectively. We evaluate \xvista from three complementary perspectives. 
First, we assess both general and embodied reasoning capabilities by comparing the two model tracks against scale-matched baselines on a comprehensive suite of benchmarks, including general vision-language understanding as well as spatial, temporal, action guidance, and state reasoning.
Second, we conduct a controlled consolidation analysis comparing the pretrained backbone (i.e., \qwenlarge), four standalone specialists, and the TIES, MOPD-only, and TIES+MOPD variants to measure capability retention and consolidation effectiveness. 
Third, beyond isolated benchmarks, we evaluate \xvista in simulated embodied environments, including EmbodiedBench~\citep{yang2025embodiedbench} and \vigil~\citep{vigil2026}, to assess how diverse reasoning capabilities compose during task execution.
These three perspectives respectively assess broad embodied reasoning, specialist capability retention, and closed-loop capability composition.

In summary, our main contributions are:\begin{itemize}[leftmargin=1.4em,itemsep=0.3em,topsep=0.25em]
  \item We introduce an execution-centric capability taxonomy comprising Spatial Reasoning, Temporal Understanding, Action Guidance, and State Verification, together with typed, verifiable output formats and same-origin capability specialists for measuring capability acquisition and retention.
  \item We present \xvista, a unified embodied reasoning model family with two model tracks, \xvistatwo and \xvistalarge, that consolidates capability specialists into a single inference-time model through TIES initialization followed by routed MOPD.
   \item We provide a multi-level evaluation spanning benchmark suites---including Capek-StateBench, a new benchmark for physical- and task-state verification---controlled capability-retention analysis, and simulated closed-loop environments, enabling separate analysis of capability acquisition, consolidation, and composition. Against their protocol-matched Qwen initializations, \xvistalarge and \xvistatwo improve 28 of 34 and 30 of 34 matched benchmark rows, respectively.
\end{itemize}

\FloatBarrier

\section{Overview}

\label{sec:model}

\subsection{Backbones and Unified Output Interface}

\xvista builds on the Qwen vision-language series~\citep{qwen36_35b_a3b, qwen3.5} and comprises two scales: 35B-A3B and 2B. Both models adopt the standard ViT--LLM composition with a vision transformer and a language-model decoder, providing general visual-language understanding and reasoning out of the box. The two tracks expose the same multimodal autoregressive interface and follow the same post-training formulation, while using independently initialized dense and MoE backbones at different capacity scales. 
Our primary model, \xvistalarge, inherits the Mixture-of-Experts (MoE) architecture of \qwenlarge: of its 35B total parameters, only about 3B are activated per token at inference. Sparse activation reduces the number of language-model parameters evaluated per token relative to a dense model with the same total parameter count.
\xvistatwo, a 2B dense model, targets resource-constrained deployment and serves as a second track for validating that the same construction transfers across capacity regimes.

All tasks are cast as text generation with no task-specific decoding heads. Points, boxes, and trajectories are emitted as normalized coordinates, temporal spans as start--end times, and state judgments as structured fields, each following the output format of its capability family (Section~\ref{sec:data}).

\subsection{Method Overview: Specialist-to-Unified Post-Training}
Because the four capability families differ in their output formats and reward geometries, we separate capability acquisition from capability integration. Within each model track, four same-origin specialists are independently post-trained under capability-aligned formats. Their task vectors are then composed with TIES to initialize a unified student, which is further consolidated through routed MOPD on student-generated prefixes. The specialist routes are used only during training; inference uses a single autoregressive checkpoint. This construction makes both capability production and capability retention auditable: the standalone experts define what should be preserved, and the consolidated model is evaluated against them.


\section{Capability-Oriented Data Construction}
\label{sec:data}

Our data construction operationalizes the execution-centric taxonomy
through four capability-aligned supervision families: \textbf{Spatial Reasoning} for geometry and relations, \textbf{Temporal Understanding} for events and changes over time,
\textbf{Action Guidance} for task referents, affordances, and trajectories, and \textbf{State Verification} for physical predicates and task states. Together, these families cover recurring information needs across a broad range of embodied tasks.

\begin{figure}[t]
\centering
\includegraphics[width=0.95\linewidth]{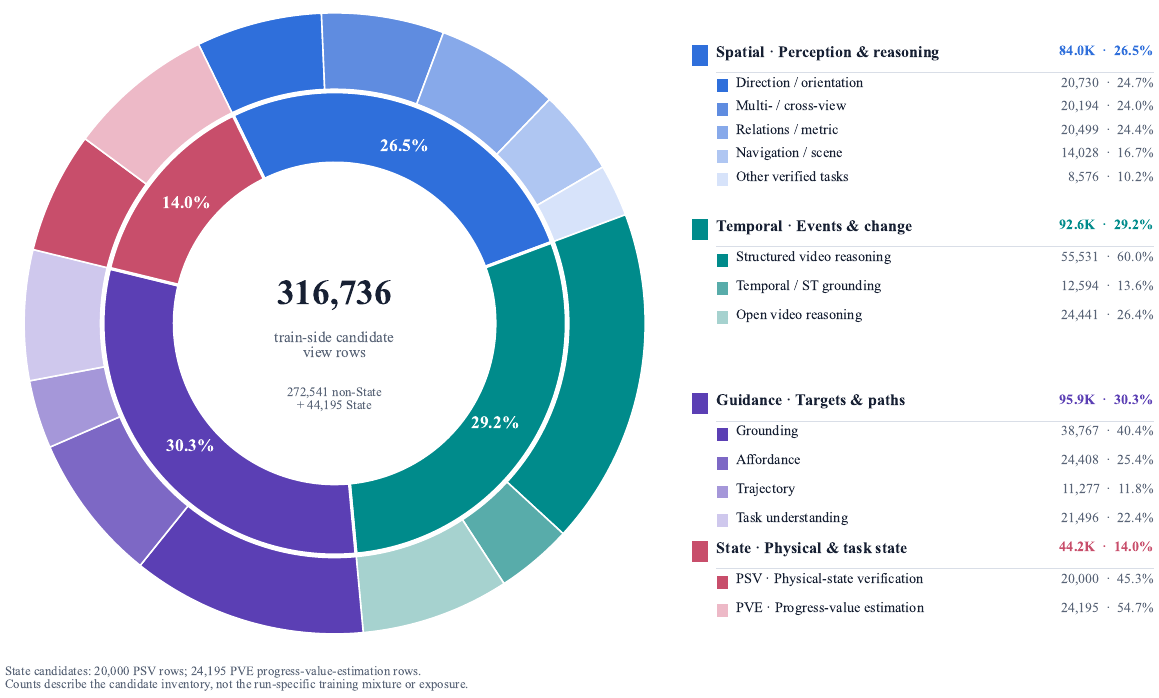} 
\caption{\textbf{Audited candidate-data snapshot.}
The inner ring groups the four capability families, and the outer ring separates their principal supervision products. All slices are reported in train-side candidate view rows.}
\label{fig:data_composition}
\end{figure}

\begin{figure}[t]
\centering
\includegraphics[width=\linewidth]{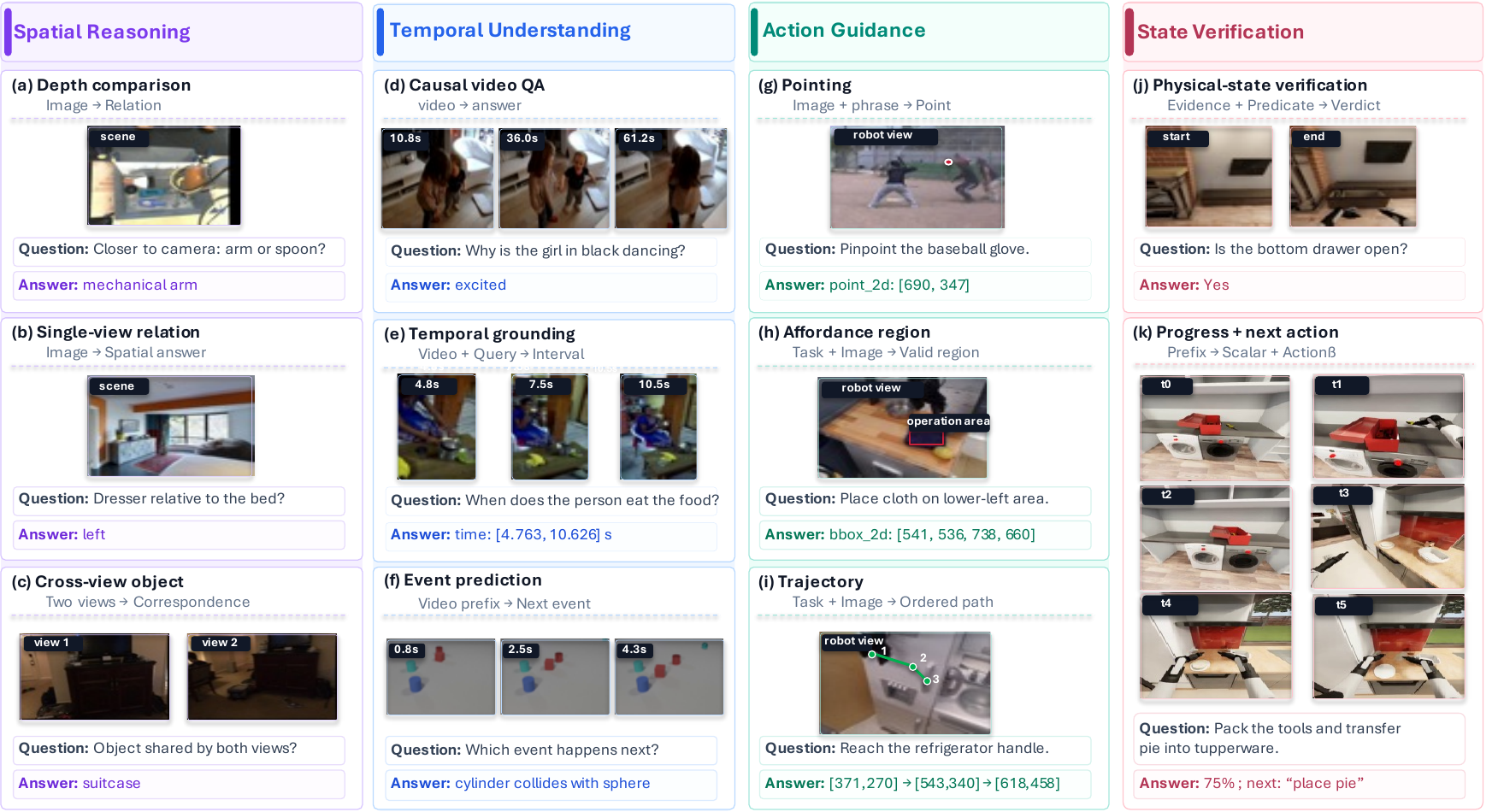}
\caption{\textbf{Representative capability-aligned supervision.} Each panel shows a source-traceable input paired with its reference target, rather than a model prediction. The examples cover spatial relation and metric reasoning, temporal reasoning and localization, action-oriented points, regions, and trajectories, as well as physical-state, task-state, and progress records. These examples are grouped into four specialist branches according to their primary execution roles; multiple output formats within one family do not define separate specialists.}
\label{fig:data_examples}
\end{figure}

Rather than grouping records solely by source dataset, we organize each record according to its primary execution role and output format. Each family provides supervision to one capability specialist. Multiple formats within the same family---for example, pointing, affordance localization, and trajectory prediction within Action Guidance---serve as complementary output forms rather than defining separate specialist branches. Standardization harmonizes serialization across families while preserving each specialist's native target semantics and verification criteria. Figure~\ref{fig:data_composition} summarizes the audited train-side candidate inventory, while Figure~\ref{fig:data_examples} illustrates representative input--target pairs.

\subsection{Spatial Reasoning}
Spatial reasoning concerns the perception of scene geometry and the
relations among entities, the observer, and different viewpoints~\citep{liu2025spatial}. It supports embodied decision-making, including
determining where one object lies relative to another, estimating target
distance and scale, assessing placement feasibility, and reasoning about how spatial relations transform across viewpoints. Our Spatial supervision combines single-view geometry with reasoning across observations, covering direction and orientation, depth and distance, relative size, support and containment, placement, and cross-view correspondence.

\smallskip\noindent\textbf{Dataset.} 
We construct the Spatial post-training corpus from
SenseNova-SI-8M~\citep{cai2026scaling},
VSI-590K~\citep{yang2025thinkinginspace},
MindCube~\citep{wang2025mindcube}, and
EmbSpatial-SFT~\citep{du2024embspatial}. These sources provide
complementary annotations at the scene and viewpoint levels. 
The resulting dataset spans four categories: (i) \emph{spatial relations}, including relative direction, orientation, and cardinal-relation reasoning; (ii) \emph{metric reasoning}, including depth, distance, and size estimation; (iii) \emph{multi-view reasoning}, including object correspondence and viewpoint transformation; and (iv) \emph{spatial scene reasoning}, including object localization,
counting, and navigation-relevant questions whose targets remain spatial
relations or scene-level judgments rather than executable action paths. Before post-training, we filter records for annotation validity, visual
support, target consistency, and duplication, and then convert the retained samples into format-specific supervision. Difficulty-aware rollout filtering is applied later during specialist RL rather than treated as a static property of the corpus.

\subsection{Temporal Understanding}
\label{sec:temporal_data}
Temporal understanding enables an embodied agent to reason about dynamic visual evidence: what happens, in what order, when it occurs, and how the scene changes over time. Unlike static-image understanding, robot execution requires the model to track event progression, causal relationships, temporal boundaries, and action-conditioned changes across observations. Accordingly, our Temporal supervision combines video reasoning with
temporal localization, requiring the model both to interpret events distributed over time and to identify when they occur.

\smallskip\noindent\textbf{Dataset.}
The temporal data covers three complementary forms of supervision.
(i) \emph{Closed-form video reasoning} develops temporal reasoning through questions about actions, object interactions, event ordering, counting, and causal relationships. We curate these samples from CLEVRER~\citep{yi2019clevrer}, NExT-QA~\citep{xiao2021nextqa}, PerceptionTest~\citep{patraucean2023perception}, STAR~\citep{wu2024star}, LLaVA-Video~\citep{zhang2024llavavideo}, and LongVideoReason~\citep{chen2025longvila-r1}, covering both short clips and long-form videos.
(ii) \emph{Temporal grounding} trains the model to localize when an event occurs by predicting an explicit temporal interval $[t_{\mathrm{start}},t_{\mathrm{end}}]$. We collect grounding annotations from Charades~\citep{sigurdsson2016charades}, DiDeMo~\citep{hendricks2017didemo}, HiREST~\citep{zala2023hierarchical}, QuerYD~\citep{oncescu2021queryd}, and LLaVA-ST~\citep{li2025llavast}. Some samples additionally include spatial localization, enabling joint spatial-temporal grounding. For these records, temporal localization remains the primary target,
while frame-level regions provide an auxiliary spatial component.
(iii) \emph{General video understanding} provides richer execution semantics in open-domain settings, including robot manipulation, egocentric interaction, and general video description. These samples are collected from ShareGPT4Video~\citep{chen2024sharegpt4video}, VideoEspresso~\citep{han2025videoespresso}, and Cosmos-Reason1 variants built on robot-video corpora~\citep{nvidia2025cosmosreason1}. We remove samples with missing timestamps, inverted or out-of-range
intervals, inconsistent event boundaries, or missing spatial annotations required for spatiotemporal reasoning.



\subsection{Action Guidance}
\label{sec:guidance_data}

Action Guidance connects language instructions to actionable visual targets that support downstream execution. The derived questions are produced through deterministic transformations, and their targets are recomputed from the underlying geometric annotations. We therefore unify visual grounding, pointing, affordance localization, and trajectory prediction within a single capability family because they all produce action-oriented visual guidance despite differing prediction formats.

\smallskip\noindent\textbf{Dataset.}
We construct the Guidance supervision from four complementary task groups. (i) \emph{Visual grounding} predicts object regions referred to by natural-language expressions. (ii) \emph{Pointing and affordance localization} identify task-relevant interaction points or regions, including explicit negative cases in which no valid target exists. (iii) \emph{Trajectory prediction} produces ordered image-space waypoint sequences for manipulation or navigation. (iv) \emph{Embodied instruction grounding} links task language to the
relevant referent, action type, interaction region, or immediate visual target.

The data are curated from PixMo~\citep{deitke2024molmo}, RoboPoint~\citep{robopoint2024}, DROID~\citep{khazatsky2024droid}, AgiBot World~\citep{agibot2025colosseo}, RoboMIND 2.0~\citep{hou2025robomind2}, FSD~\citep{yuan2025fsd}, ShareRobot~\citep{robobrain2025}, RoboVQA~\citep{sermanet2023robovqa}, Robo2VLM~\citep{chen2025robo2vlm}, RoboAfford++~\citep{hao2025roboafford}, SPAR~\citep{zhang2025spar}, and OneThinker~\citep{feng2025onethinker}, covering manipulation, navigation, embodied question answering, and robot instruction following.
To enable unified post-training, all coordinate-based targets are converted from their native annotations to a shared $[0,1000]^2$ image-coordinate system. Point, box, and affordance annotations preserve their original target semantics, including explicit negative cases and multiple valid interaction locations where applicable. Trajectory annotations retain ordered waypoint sequences rather than unordered point sets; navigation trajectories additionally preserve sparse semantic waypoints and, when available, denser 2D paths. This representation preserves trajectory order, endpoints, and path geometry for downstream verification while standardizing the supervision format across guidance tasks. Representative examples are shown in Figure~\ref{fig:data_examples}.

\begin{figure}
    \centering
    \includegraphics[
  width=\textwidth
]{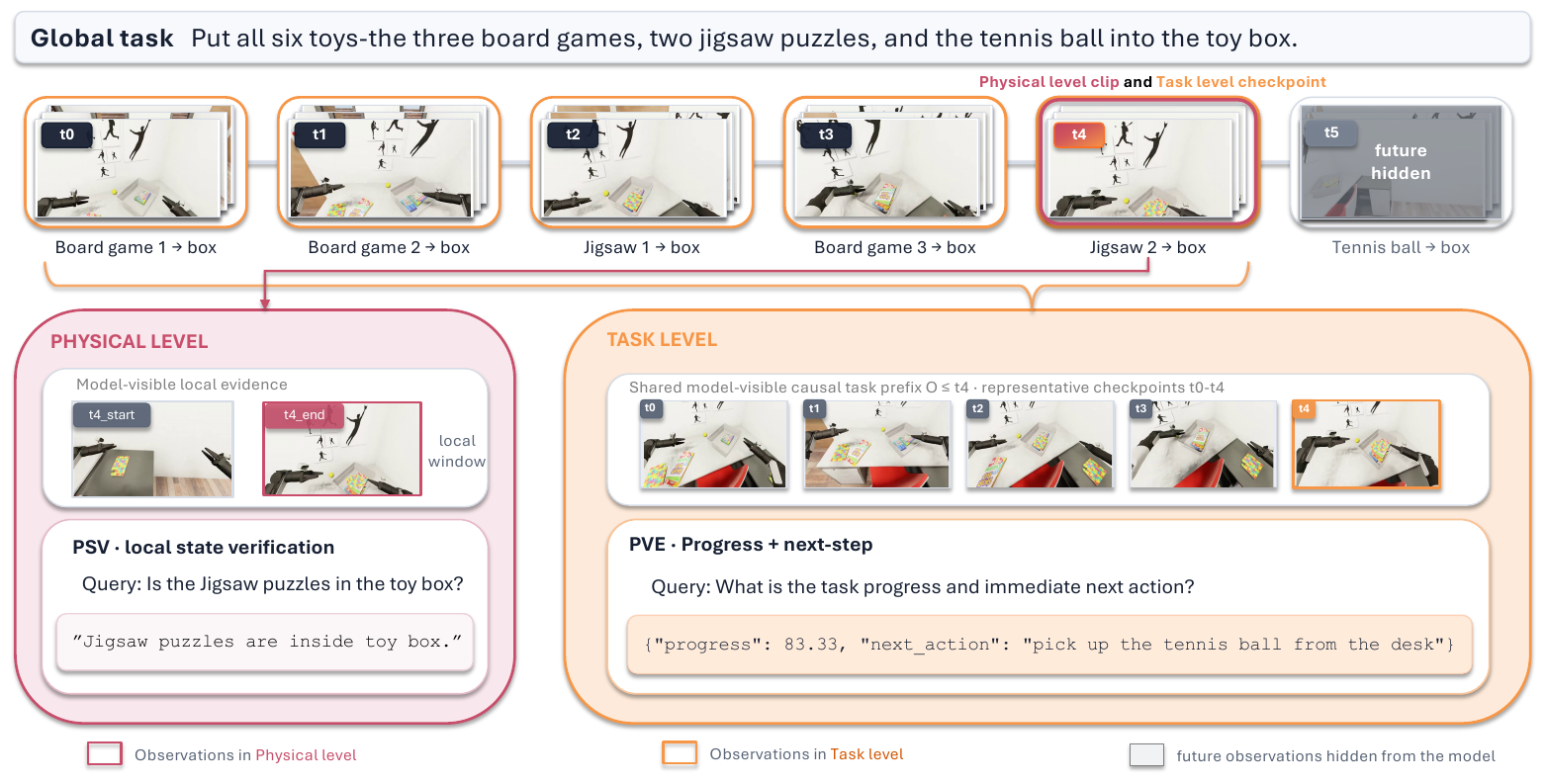}
    \caption{\textbf{Execution-centric construction of State Verification supervision.}
Starting from one BEHAVIOR-1K trajectory and its global task instruction, we retain checkpoint-aligned evidence segments at two semantic levels. Physical-state verification uses a local transition window to verify an action-result predicate, whereas task-level supervision uses the causal observation prefix through the current checkpoint. The shared task prefix supports both explicit goal-condition verification and the compact progress-value estimation readout. Future observations are hidden from the model.}
    \label{fig:state_data_construction}
\end{figure}

\subsection{State Verification}

State verification assesses whether the visually observed execution state supports the conditions required to continue or terminate a robot task. We formulate this capability at two execution-aligned levels. At the \emph{physical level}, the model determines whether a recent action has produced the required object state or relation. At the \emph{task level}, it estimates how much of the high-level instruction has been completed and what should happen next.

\smallskip\noindent\textbf{Dataset.}
We construct all State Verification post-training data from BEHAVIOR-1K household-manipulation trajectories~\cite{li2023behavior1k}. Rather than sampling observations uniformly, we anchor examples at execution-relevant primitive, skill, and task checkpoints. Task hierarchies, transition records, official BDDL goal conditions, simulator states, action histories, and future observations may be used on the teacher side to select checkpoints and derive or audit supervision targets. The model receives only the instruction and the causal visual evidence available at the selected checkpoint.

We materialize two model-facing supervision types. At the \emph{physical level}, \emph{physical-state verification (PSV)} uses minimal local causal evidence to determine whether an object state or relation satisfies the requirement of the current primitive or skill. The retained evidence may be a single critical frame, an aligned frame list, or a short transition clip, and covers states and relations such as open or closed, powered on or off, grasp and release, containment, support, and other action-result predicates.

At the \emph{task level}, \emph{progress-value estimation (PVE)} uses the causal visual task prefix to estimate how much of the instruction has been completed and, for paired examples, predict the immediate annotated next action. Each record explicitly identifies whether progress is defined by the fraction of satisfied BEHAVIOR-1K goal conditions or by the fraction of completed primitive child skills, including navigation; the two progress definitions remain separate label spaces. Progress-only and progress-plus-next-step records are treated as output variants of the same PVE supervision type. The underlying goal-condition or child-skill decomposition is used to define these targets but is not itself exposed as an output. In this way, PVE ties progress and next-step supervision to the component structure of the task without exposing symbolic execution traces or future observations.

Before training, samples are filtered for visual support, annotation consistency, and duplication. Policy-dependent difficulty filtering is applied later during specialist RL rather than treated as a static property of the corpus. Figure~\ref{fig:state_data_construction} illustrates how a single trajectory is materialized into local physical-state evidence and shared causal task-state evidence.

\subsection{Data Standardization}

After capability-specific construction, all samples are converted into a unified record format while preserving their original supervision targets. 
Every sample undergoes automatic validation to ensure media integrity, annotation consistency, and format correctness. Depending on the task, additional semantic checks verify that supervision targets remain visually supported after preprocessing.
Dataset splits are inherited from the original sources and assigned before view serialization so that all aligned views of the same logical sample remain in the same split. Duplicate removal is performed at the logical-sample level to avoid leakage across training and evaluation.

\section{Post-Training}
\label{sec:recipe}

\begin{figure}[t]
  \centering
  \includegraphics[width=0.99\linewidth]{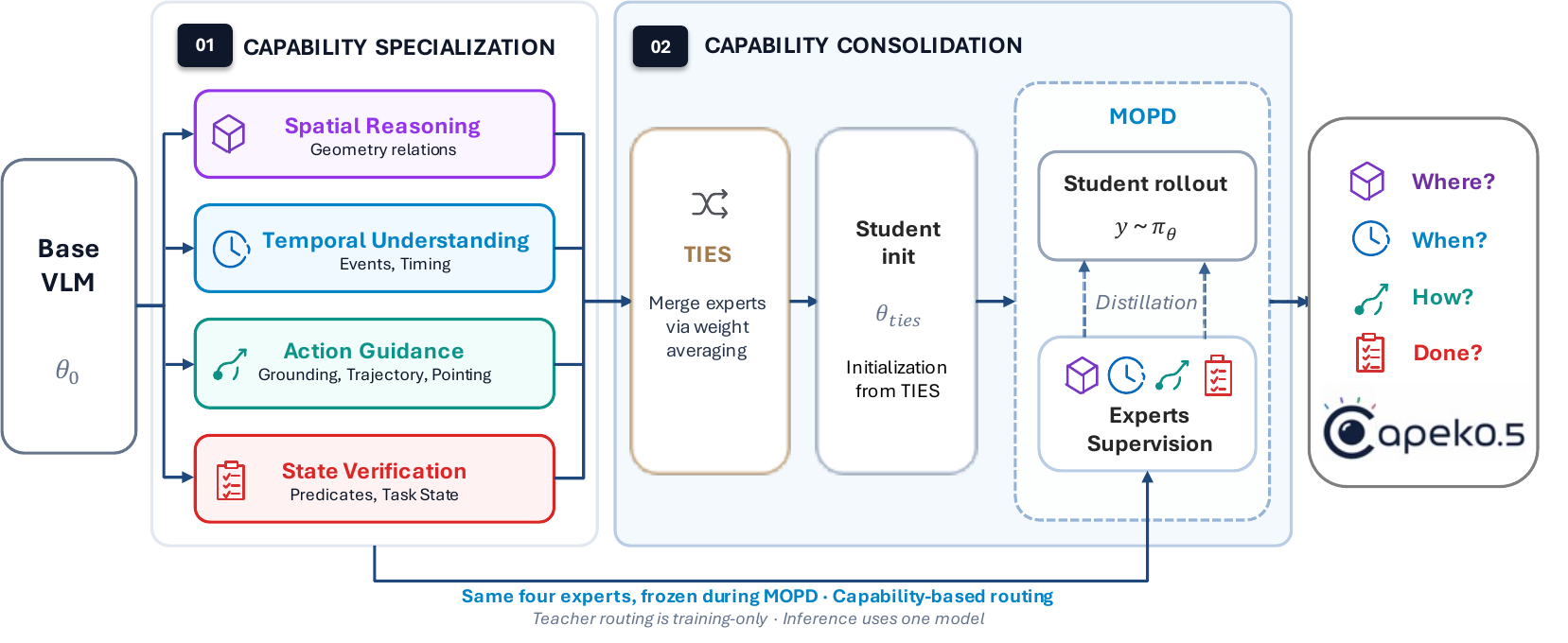}
  \caption{\textbf{Capability specialization and consolidation.} Four specialists acquire complementary execution-facing capabilities from a shared backbone. TIES combines their task vectors to initialize a unified student, after which routed MOPD distills the responsible specialist on prefixes induced by that student. Specialists and routing are required only during consolidation; deployment uses one checkpoint.}
  \label{fig:post_training_pipeline}
\end{figure}

\xvista adopts a two-stage post-training paradigm that separates capability acquisition from capability consolidation.
In the first stage, four capability specialists are independently optimized from the same backbone checkpoint using capability-specific data, objectives, and rewards.
In the second stage, the specialists are consolidated into a single model through weight-space merging followed by on-policy distillation. The specialist teachers and routing mechanism are used only for training; inference requires only a single autoregressive model.
Figure~\ref{fig:post_training_pipeline} summarizes the training paradigm:

\begin{itemize}[leftmargin=1.35em,labelsep=0.5em]
    \item \textbf{Capability Specialization.}
    The Spatial Reasoning, Temporal Understanding, Action Guidance, and State Verification specialists are trained independently, each with its own data and rewards.

    \item \textbf{Capability Consolidation.}
    The specialists are first merged with TIES to initialize a unified student, which is then refined via routed Multi-Teacher On-Policy Distillation (MOPD) that transfers their complementary behaviors into a single policy.
\end{itemize}

\subsection{Capability Specialist Training}

In the capability specialization stage, all specialists are independently optimized from a shared backbone with identical architecture and tokenizer. Despite their heterogeneous output formats, they are trained under a unified reward formulation with task-specific parsers and verifiers using Group Relative Policy Optimization (GRPO), which keeps the resulting checkpoints parameter-compatible for consolidation.

\subsubsection{Token-Level GRPO Optimization}

All specialists are optimized with GRPO~\citep{deepseekmath2024,guo2025deepseekr1}. For a multimodal input--target pair $(x,y^\star)$, the rollout policy $\pi_{\theta_{\mathrm{old}}}$ samples a group of $G$ responses $\{y_i\}_{i=1}^{G}$, each scored by its format-specific reward $r_i=R(y_i,y^\star)$. GRPO computes a response-level advantage by normalizing rewards within the group,
\begin{equation}
\hat A_i=\frac{r_i-\bar r}{\sigma_r+\delta},
\label{eq:group_advantage}
\end{equation}
where $\bar r=G^{-1}\sum_i r_i$ and $\sigma_r$ are the group mean and standard deviation of the rewards and $\delta>0$ is a numerical stabilizer. This group-relative baseline removes the need for a learned value function. The advantage $\hat A_i$ is assigned to every token of $y_i$, and the policy is updated by the token-level clipped objective
\begin{equation}
\mathcal{L}_{\mathrm{RL}}(\theta)
=
-\frac{1}{\sum_{i=1}^{G}|y_i|}
\sum_{i=1}^{G}\sum_{t=1}^{|y_i|}
\Big[
\min\!\big(
\rho_{i,t}\,\hat A_i,\;
\operatorname{clip}(\rho_{i,t},1-\epsilon,1+\epsilon)\,\hat A_i
\big)
-\beta\,\widehat D_{\mathrm{KL},i,t}
\Big],
\label{eq:grpo}
\end{equation}
where $
\rho_{i,t}=\frac{\pi_\theta(y_{i,t}\mid x,y_{i,<t})}{\pi_{\theta_{\mathrm{old}}}(y_{i,t}\mid x,y_{i,<t})}
$
is the token-level importance ratio, $\epsilon$ is the clipping threshold, $\beta\geq0$ is the KL coefficient, and $\widehat D_{\mathrm{KL},i,t}$ is the KL regularization.

\smallskip\noindent\textbf{Rollout-based sample selection.}
The advantage in Equation~\ref{eq:group_advantage} vanishes when all responses to a prompt receive the same reward. We therefore prioritize prompts with nonzero within-group reward variance: uniformly high-reward prompts are downsampled, while uniformly low-reward prompts are audited for annotation ambiguity, insufficient visual evidence, parser failures, or genuine difficulty before being retained or removed. Screening is performed independently per capability, since parsers and reward geometries differ across output formats.

\FloatBarrier

\begin{table}[t]
\centering
\scriptsize
\setlength{\tabcolsep}{3.4pt}
\renewcommand{\arraystretch}{1.13}
\caption{\textbf{Reward formulation for different output formats.} Each output format combines binary format validity with one or more task-specific reward functions. The same scoring functions are reused whenever identical output types appear across specialists.}
\label{tab:rewards}
\begin{tabularx}{0.99\linewidth}{
    >{\raggedright\arraybackslash}p{0.15\linewidth}
    >{\raggedright\arraybackslash}p{0.18\linewidth}
    >{\raggedright\arraybackslash}p{0.21\linewidth}
    >{\raggedright\arraybackslash}X}
\toprule
\textbf{Reward term} & \textbf{Specialist(s)} & \textbf{Applied format(s)} & \textbf{Scoring rule} \\
\midrule
Format validity
  & All four
  & Every declared output format
  & Binary success of format-specific hard parsing \\
\rowcolor{gray!8}
Answer correctness
  & Spatial Reasoning, Temporal Understanding
  & Choice, set, Boolean, scalar, OCR, and QA
  & Exact or set agreement, numerical tolerance, edit similarity, or frozen semantic score \\
Temporal overlap
  & Temporal Understanding
  & Event interval and space--time localization
  & Temporal IoU; space--time formats average temporal and frame-level box overlap \\
\rowcolor{gray!8}
Box overlap
  & Temporal Understanding, Action Guidance
  & Frame-level space--time boxes and grounded box sets
  & Frame-averaged or one-to-one 2D IoU, with missing and unmatched items contributing zero \\
Point / region / affordance
  & Action Guidance
  & Pointing, region, affordance, and no-target formats
  & Point F1, target-mask hit, symmetric nearest-neighbor score, or exact abstention \\
\rowcolor{gray!8}
Trajectory agreement
  & Action Guidance
  & Ordered waypoint sequence
  & Exponential reward over raw DFD after coordinate normalization and five-point arc-length resampling \\
Physical-state correctness
  & State Verification
  & Direct-state selection and binary predicate verification
  & Normalized exact agreement with the target state token or \texttt{yes}/\texttt{no} verdict \\
\rowcolor{gray!8}
Progress and next-step agreement
  & State Verification
  & Progress-and-next-action JSON
  & Truncated progress proximity, normalized next-action agreement, and binary JSON validity \\
\bottomrule
\end{tabularx}
\end{table}

\subsubsection{Capability-Specific Reward Design}
\label{sec:reward_design}

Although the four specialists produce heterogeneous outputs, they all share the same reward structure: an optional format reward and a task-accuracy reward. The format reward checks whether the output is well-formed and parseable (e.g., a bounding box has four ordered coordinates) while the accuracy reward measures correctness against the target. The total reward is their weighted sum,
\begin{equation}
R(\hat y,y^\star)
=
\lambda_{\mathrm{fmt}}\,R_{\mathrm{fmt}}(\hat y)
+
\lambda_{\mathrm{acc}}\,R_{\mathrm{acc}}(\hat y,y^\star),
\label{eq:format_reward}
\end{equation}
where $R_{\mathrm{fmt}}\in\{0,1\}$ is the format reward and $R_{\mathrm{acc}}\in[0,1]$ is the accuracy score, computed by a rule-based verifier or, for open-ended outputs, a frozen semantic judge. The nonnegative weights sum to one, so $R\in[0,1]$. Table~\ref{tab:rewards} lists the output formats and their scoring rules.

\smallskip\noindent\textbf{Spatial Reasoning.}
Spatial reasoning tasks span relational understanding, metric estimation, cross-view correspondence, and viewpoint reasoning. Their outputs primarily follow standard verifiable reasoning formats, including multiple-choice, Boolean, set-valued, and numerical answers. Accordingly, rewards are computed using exact or set agreement, numerical tolerance, or one-to-one correspondence matching according to the output type.

\smallskip\noindent\textbf{Temporal Understanding.}
The Temporal Understanding specialist covers video understanding, event ordering, temporal localization, and space--time grounding. Categorical reasoning tasks reuse the scoring rules from Spatial Reasoning. For temporal localization, a valid predicted interval $I_p=[s_p,e_p]$ is compared with the reference interval $I_g=[s_g,e_g]$ using temporal Intersection-over-Union (tIoU)~\citep{hendricks2017didemo}:
\begin{equation}
R_{\mathrm{tIoU}}(I_p,I_g)
=
\frac{\max\!\left(0,\min(e_p,e_g)-\max(s_p,s_g)\right)}
{\max(e_p,e_g)-\min(s_p,s_g)}.
\label{eq:temporal_iou}
\end{equation}
Predictions that fail format validation receive zero reward. For space--time grounding, temporal and spatial evidence are weighted equally:
$
R_{\mathrm{st}}
=
\tfrac12R_{\mathrm{tIoU}}
+
\tfrac12\overline{R}_{\mathrm{box}},
$
where $\overline{R}_{\mathrm{box}}$ denotes the mean box IoU over the annotated frames, with missing boxes contributing zero.

\smallskip\noindent\textbf{Action Guidance.}
The Action Guidance specialist covers visual grounding, pointing, affordance localization, and trajectory prediction~\citep{lu2026visualprimitives,embodiedr12025}. Set-valued grounding tasks use one-to-one matching, with unmatched predictions or targets contributing zero. Region and affordance predictions are evaluated using their corresponding mask- or distance-based metrics, while explicit no-target cases reward correct abstention.

Ordered trajectories are evaluated using the discrete Fr\'echet distance (DFD), which preserves waypoint order~\citep{eiter1994computing}. Coordinates are first mapped to $[0,1]^2$, after which both predicted and reference paths are resampled at five uniformly spaced arc-length positions, yielding $\bar P$ and $\bar Q$. We use raw DFD as the geometric discrepancy and convert it to a higher-is-better reward with an exponential kernel:
\begin{equation}
C_{\mathrm{traj}}(\bar P,\bar Q)
=
D_{\mathrm{DFD}}(\bar P,\bar Q),
\qquad
R_{\mathrm{traj}}(\bar P,\bar Q)
=
\exp\!\left(
-\lambda_{\mathrm{traj}}
C_{\mathrm{traj}}(\bar P,\bar Q)
\right),
\quad
\lambda_{\mathrm{traj}}=10.
\label{eq:trajectory_reward}
\end{equation}
For valid trajectories, the reward lies in $(0,1]$, with exact matches receiving one and larger geometric discrepancies receiving exponentially smaller rewards. Invalid or empty trajectories receive zero reward through format validation. Appendix~\ref{app:trajectory_scoring} provides the complete DFD definition and details.
    
\smallskip\noindent\textbf{State Verification.}
The State Verification specialist comprises two supervision types: Physical-State Verification (PSV) and Progress-Value Estimation (PVE). At the physical level, PSV predicts normalized object states or binary physical predicates from visual observations, and rewards exact agreement with the reference state or \texttt{yes}/\texttt{no} verdict.
At the task level, PVE estimates execution progress from the observed execution history and, when available, predicts the annotated immediate next action. Progress supervision follows the goal-condition and child-skill decompositions described in Section~\ref{sec:data}. 

Given predicted and reference progress values $\hat p,p\in[0,100]$, the progress reward is defined as a 25-percentage-point truncated proximity score:
$
r_{\mathrm{value}}
=
\max\!\left(0,1-\frac{|\hat p-p|}{25}\right).
$
For samples with next-action supervision, the overall reward is
$
R_{\mathrm{prog+act}}
=
0.5r_{\mathrm{value}}
+
0.4r_{\mathrm{action}}
+
0.1r_{\mathrm{fmt}}
$,
where $r_{\mathrm{action}}$ denotes normalized agreement with the annotated next action (including \texttt{null} for completed tasks), and $r_{\mathrm{fmt}}$ indicates JSON validity. Samples without next-action supervision are optimized using only $r_{\mathrm{value}}$ after the same format validation.

\subsection{Capability Consolidation}
After capability specialization, the four experts are consolidated into a single model through a two-step procedure. We first perform weight-space merging using TIES to obtain a unified initialization, which combines complementary parameter updates while mitigating conflicting weight changes. Starting from this initialization, routed Multi-Teacher On-Policy Distillation (MOPD) further transfers the specialists' behaviors into a single policy by supervising student-generated trajectories.

\subsubsection{Weight-Space Initialization}
\label{sec:ties_initialization}

TIES provides a conflict-aware weight-space initialization for the unified student. Concretely,
let $\theta_0$ be the shared backbone and $\theta_e$ the checkpoint of specialist $e$. Since all specialists share the same architecture, tokenizer, and initialization, their parameter updates are directly compatible:
$\Delta_e=\theta_e-\theta_0$. We use TIES~\citep{ties2023,ilharco2023taskarithmetic}, which resolves conflicts among specialist updates by trimming low-magnitude parameters, electing a coordinate-wise consensus sign, and averaging only sign-consistent updates:
\begin{equation}
\theta_{\mathrm{TIES}}
=
\theta_0
+
\lambda\,
\operatorname{TIES}_{\tau}\!\left(
\{\Delta_e\}_{e=1}^{E}
\right),
\label{eq:ties_initialization}
\end{equation}
where $E=4$, $\tau=0.8$ masks the $80\%$ lowest-magnitude entries within each task vector, and $\lambda=1.0$ is the merged-update scale. The resulting checkpoint $\theta_{\mathrm{TIES}}$ initializes the unified student for routed MOPD.

\subsubsection{Routed MOPD}
\label{sec:mopd_consolidation}
Parameter merging alone does not guarantee that every specialist behavior is preserved in the merged model. We therefore further refine the TIES-initialized student using routed MOPD~\citep{mopd2026,gkd2023}. During training, each sample is routed to its corresponding specialist teacher, which supervises the student on its own generated trajectories. This allows each capability to be distilled directly from the responsible specialist.

Concretely, for an input $x$, the student first samples a response $y\sim\pi_{\theta}(\cdot\mid x)$, and the routed teacher $q_r$ is then evaluated on the resulting student-generated prefixes $h_t=(x,y_{<t})$, where $r$ indexes the capability route selected for the sample. The student is optimized to minimize the token-level reverse KL to the routed teacher:
\begin{equation}
\mathcal L_{\mathrm{MOPD}}(\theta)
=
\mathbb E\!\left[
\frac{1}{|y|}\sum_{t}
D_{\mathrm{KL}}\!\left(
\pi_\theta(\cdot\mid h_t)\,\|\,q_r(\cdot\mid h_t)
\right)
\right].
\label{eq:mopd_reverse_kl}
\end{equation}
Throughout training, only the routed teacher is queried for each sample, and all specialist teachers remain frozen. In practice, rather than evaluating the full-vocabulary KL, we optimize a sampled surrogate: the teacher supplies the log-probability of each student-sampled token, which defines a per-token distillation advantage
\begin{equation}
\hat A_t^{\mathrm{MOPD}}
=
\operatorname{clip}\!\left(
\operatorname{sg}\!\left[
\log\frac{q_r(y_t\mid h_t)}{\pi_\theta(y_t\mid h_t)}
\right],
-\epsilon_{\max},\,\epsilon_{\max}
\right),
\label{eq:mopd_advantage}
\end{equation}
where $\operatorname{sg}[\cdot]$ is the stop-gradient operator and $\epsilon_{\max}>0$ clips extreme advantages to stabilize training. 
Since rollouts are generated on-policy from $\pi_\theta$, this advantage is optimized directly with the token-mean policy-gradient objective.

For the controlled ablation in Section~\ref{sec:expert_integration}, we additionally apply MOPD directly to the backbone checkpoint to isolate the effect of weight-space initialization and conduct comparison analysis. Both variants require only a single student checkpoint at inference.

\section{Experiments}
\label{sec:experiments}

We evaluate both \xvistalarge and \xvistatwo on a comprehensive suite of general and embodied benchmarks spanning Spatial Reasoning, Temporal Understanding, Action Guidance, and State Verification. We then conduct a controlled consolidation study by comparing the four capability specialists, the Mix-RL checkpoint, the TIES initialization, the MOPD-only student, and the released TIES+MOPD model to analyze capability retention throughout consolidation. Finally, we evaluate the unified model in closed-loop embodied interaction using EmbodiedBench~\citep{yang2025embodiedbench} and \vigil~\citep{vigil2026}.

\subsection{Experimental Setup}
\label{sec:evaluation_design}

\subsubsection{Capek-StateBench}
\label{sec:statebench}

Existing public benchmarks provide broad coverage of spatial reasoning, temporal understanding, and action guidance, but they rarely evaluate whether a model can verify the resulting world state after execution or assess task progress toward completion. To evaluate this capability, we introduce \textbf{Capek-StateBench}, a benchmark for State Verification consisting of two complementary tracks: \textbf{Capek-StateBench-P} for physical-state verification and \textbf{Capek-StateBench-T} for task-state verification.

\smallskip\noindent\textbf{Physical-State Verification (Capek-StateBench-P).}
The physical track evaluates whether a model can determine observable object states and spatial relations at annotated execution boundaries. Each sample consists of a single image, a short video clip, or a small set of key frames, together with a verification query. Queries either require selecting the correct state from predefined alternatives or verifying a physical predicate using a binary \texttt{yes}/\texttt{no} response.

\smallskip\noindent\textbf{Task-State Verification (Capek-StateBench-T).}
The task track evaluates a model's understanding of execution progress from a causal visual prefix. Given a high-level task instruction and the observed execution history, the model predicts (1) the current task progress and (2) the immediate next action required to continue execution; this track therefore instantiates the execution-level progress-and-next-step readout used during State Verification training (Section~\ref{sec:recipe}). Progress is defined using two complementary annotation schemes: task-condition progress measures the fraction of satisfied BEHAVIOR-1K~\citep{li2023behavior1k} goal conditions, while primitive-skill progress measures the completion ratio of annotated low-level skills, including navigation. The released 500-example split combines both schemes, with 213 task-condition records and 287 primitive-skill records. The primary score is computed over the combined split, while annotation-type breakdowns are retained as diagnostics; Appendix~\ref{app:statebench_protocol} gives the complete protocol.

\smallskip\noindent\textbf{Evaluation Metrics.}
For the physical track, we report normalized exact-match accuracy together with breakdowns by query type, evidence source, state category, and execution level. For the task track, progress is scored using
\begin{equation}
s_p=\max\!\left(0,1-\frac{|\hat p-p|}{25}\right),
\label{eq:progress_score}
\end{equation}
while next-step prediction is evaluated by semantic exact matching, denoted by $s_a$. The primary metric combines both aspects:
\begin{equation}
s_{\mathrm{T}}=\frac{5s_p+4s_a}{9}.
\label{eq:pve_score}
\end{equation}
We additionally report progress mean absolute error, progress score, next-step accuracy, terminal versus non-terminal performance, label-space breakdowns, and output validity as diagnostic metrics. Qualitative examples from both Capek-StateBench tracks, together with the corresponding model-visible evidence and complete model rollouts, are provided in Appendix~\ref{app:static_qa_rollouts}.

\subsubsection{Benchmark Overview}

We organize the offline evaluation according to the primary capability exercised by each benchmark. Each benchmark is assigned to a single capability group based on its dominant prediction type, while retaining its original evaluation protocol and native metric.

\begin{itemize}[leftmargin=1.35em,itemsep=3pt,topsep=4pt]
\item \textbf{Spatial Reasoning.} Benchmarks cover geometric reasoning, spatial relations, viewpoint understanding, and embodied scene understanding, including CV-Bench~\citep{tong2024cambrian}, VSI-Bench~\citep{yang2025thinkinginspace}, OmniSpatial~\citep{jia2025omnispatial}, MindCube~\citep{wang2025mindcube}, RoboSpatial-Home~\citep{song2024robospatial}, All-Angles-Bench~\citep{yeh2025allangles}, EmbSpatial-Bench~\citep{du2024embspatial}, ERQA~\citep{geminirobotics2025}, and OpenEQA~\citep{majumdar2024openeqa}.

\item \textbf{Temporal Understanding.} Benchmarks cover event ordering, temporal localization, and long-video understanding, including EgoTempo~\citep{plizzari2025omnia}, Video-MME~\citep{fu2025video}, MVBench~\citep{li2024mvbench}, LongVideoBench~\citep{wu2024longvideobench}, and QVHighlights-TimeLens~\citep{zhang2025timelens}.
For evaluation, QVHighlights-TimeLens reports temporal mIoU. 

\item \textbf{Action Guidance.} Benchmarks cover visual grounding, pointing, affordance localization, placement, and trajectory prediction, where task language must be grounded into actionable image-space outputs, including RoboRefIt~\citep{lu2023vlgrasp}, PointBench~\citep{geminirobotics15}, VABench-point/trace~\citep{yuan2025fsd}, PixMoPointsEval~\citep{deitke2024molmo}, Where2Place~\citep{robopoint2024}, ShareRobot-Affordance/Trajectory~\citep{robobrain2025}, PIO~\citep{xue2025pio}, and NaviTrace~\citep{windecker2025navitrace}.
For evaluation, box grounding rows use IoU@0.5 in the benchmark coordinate system. ShareRobot-Affordance reports mean IoU over predicted and reference affordance boxes, with missing or invalid predictions counted as zero. Point rows use their benchmark-native point or mask scorers, including PixMoPointsEval point-level F1 and the PIO point-in-mask score. Trajectory rows keep their native trajectory metrics: NaviTrace reports its semantic-aware normalized score, VABench-trace reports trajectory RMSE, and ShareRobot-Trajectory reports raw discrete Fr\'echet distance.

\item \textbf{State Verification.} 
Capek-StateBench-P and Capek-StateBench-T correspond to the physical and task levels of State Verification. The Physical track tests local object states and relations and is scored by normalized exact-match accuracy. The Task track probes task-state understanding through execution-progress estimation and immediate next-step prediction; we report the deterministic progress score, next-step accuracy, and their predeclared joint score.

\item \textbf{General retention.} General multimodal and language benchmarks serve as regression controls for measuring whether general capabilities are retained after embodied post-training, including MMMU~\citep{yue2024mmmu}, MMVet~\citep{yu2023mmvet}, RealWorldQA~\citep{xai2024realworldqa}, MMBench-EN~\citep{liu2023mmbench}, IFEval~\citep{zhou2023ifeval}, MMLU-Pro~\citep{wang2024mmlupro}, BFCL-v3~\citep{patil2025bfcl}, and LiveCodeBench v6~\citep{jain2024livecodebench}. Together, they check retention of broad multimodal understanding, instruction following, reasoning, function-calling, and coding abilities.

\end{itemize}


\subsubsection{Evaluation Protocol}

We systematically evaluate \xvistalarge and \xvistatwo in Tables~\ref{tab:public_results_large} and~\ref{tab:public_results_compact}, respectively, following the capability groups defined above.
All results are produced with DeepInsight~\citep{deepinsight2026}, our evaluation harness, which records the executed configuration, generations, and scores of every run. For each benchmark, the dataset version, media sampling, prompt template, answer parser, scorer, and judge (when applicable) are identical across all compared models. Decoding configurations follow a two-level policy: external baselines use the inference settings recommended by their official papers or model cards, whereas each matched \xvista--Qwen pair additionally shares an identical decoding configuration, so that signed differences within a pair can be attributed to the post-training recipe.

We report the 35B-A3B track as the main comparison and the 2B track as a compact-capacity counterpart, using the same benchmark rows for both. Table~\ref{tab:public_results_large} presents the primary 35B-A3B results and ranks A3B MoE models together with 7--9B dense models whose active parameter counts are comparable, following the active-compute comparison principle used by HY-Embodied-VLM-1.0~\citep{hyembodied10_2026}. Table~\ref{tab:public_results_compact} then reports the 2B track against models with at most 4B parameters. The external baselines shown in the tables are Qwen~\citep{bai2025qwen3}, RynnBrain and RynnBrain1.1~\citep{rynnbrain2026}, MiMo-Embodied~\citep{mimoembodied2025}, Embodied-R1.5~\citep{embodiedr15}, RoboBrain-family checkpoints~\citep{robobrain2025,robobrain25_2026}, and HY-Embodied-VLM-1.0~\citep{hyembodied10_2026}.
All completed entries in both tables, including those for external baselines, are DeepInsight reruns; no scores are quoted from official reports.


\subsubsection{Inference Settings}
For \xvista and its shared Qwen start, the released DeepInsight configuration uses vLLM with a 128K model context, automatic truncation, at most 64 video frames, temperature 0.7, top-$p$ 0.95, top-$k$ 20, up to 16{,}384 generated tokens, and thinking enabled. The postprocessor removes \texttt{<think>} and optional \texttt{<answer>} wrappers before benchmark scoring. Tasks then apply their declared native format: for example, grounding uses IoU matching at 0.5 in the shared $[0,1000]$ coordinate system, ShareRobot-Affordance reports mean IoU, point tasks use benchmark-native point or mask matching, and trajectory tasks use the corresponding normalized score, RMSE, or raw DFD scorer. Model-judged rows share one non-thinking judge profile. DeepInsight records the executed configuration, generations, and scores for each run~\citep{deepinsight2026}. Appendix~\ref{app:deepinsight_protocol} gives the prompts and output formats used for the different benchmark types.

\subsection{Overall Benchmark Results}
\label{sec:broad_results}

\begin{table}[t]
\centering
\caption{Benchmark results for the 35B-A3B variant, compared with A3B MoE and 7--9B dense models of comparable active compute. 
}
\label{tab:public_results_large}
\begingroup
\scriptsize
\setlength{\tabcolsep}{1.6pt}
\renewcommand{\arraystretch}{1.03}
\begin{tabularx}{0.99\textwidth}{@{}>{\raggedright\arraybackslash}p{0.09\linewidth}>{\raggedright\arraybackslash}p{0.155\linewidth}>{\centering\arraybackslash}X>{\columncolor{XPLink!12}\centering\arraybackslash}X>{\centering\arraybackslash}X>{\centering\arraybackslash}X>{\centering\arraybackslash}X>{\centering\arraybackslash}X>{\centering\arraybackslash}X@{}}
\toprule
\textbf{Task family} & \textbf{Benchmark} &
\shortstack{Qwen3.6\\35B-A3B} &
\shortstack{\textcolor{XPNavy}{\textbf{Capek0.5}}\\\textcolor{XPNavy}{\textbf{35B-A3B}}} &
\shortstack{RynnBrain\\30B-A3B} &
\shortstack{MiMo-Emb.\\7B} &
\shortstack{Embodied\\R1.5-8B} &
\shortstack{RynnBrain1.1\\9B} &
\shortstack{HY-Embodied1.0\\30B-A3B} \\
\midrule
\multirow{10}{=}{\raggedright\bfseries\textcolor{XPNavy}{Spatial Reasoning}}
 & CV-Bench                  & 88.56 & \textbf{89.99} & 88.62 & 87.55 & 87.03 & 88.76 & \underline{89.39} \\
 & VSI-Bench                 & 60.83 & 70.69 & \underline{75.30} & 51.58 & 64.78 & \textbf{75.73} & 59.47 \\
 & OmniSpatial               & \textbf{58.84} & 56.36 & 51.27 & 43.84 & 47.36 & 48.08 & \underline{56.56} \\
 & MindCube                  & 58.67 & \underline{69.90} & 61.62 & 35.71 & 32.76 & \textbf{79.24} & 64.67 \\
 & RoboSpatial-Home          & 63.46 & \textbf{72.80} & 63.95 & 62.19 & 63.75 & 59.67 & \underline{65.98} \\
 & All-Angles-Bench          & \textbf{65.43} & \underline{64.12} & 50.70 & 53.85 & 50.19 & 56.19 & 62.85 \\
 & EmbSpatial-Bench          & \underline{82.94} & \textbf{84.59} & 80.99 & 77.64 & 76.62 & 82.50 & 82.25 \\
 & ERQA                      & \textbf{60.00} & \underline{58.25} & 44.00 & 43.25 & 43.00 & 47.50 & 57.25 \\
 & OpenEQA                   & \underline{67.48} & \textbf{69.51} & 56.17 & 57.24 & 55.58 & 58.45 & 58.48 \\
\cmidrule(lr){1-9}
\multirow{10}{=}{\raggedright\bfseries\textcolor{XPNavy}{Action Guidance}}
 & RoboRefIt                 & \underline{84.45} & \textbf{85.96} & 83.16 & 77.63 & 83.48 & 82.09 & 82.20 \\
 & PointBench                & \underline{76.66} & \textbf{81.22} & 53.91 & 58.01 & 61.72 & 53.63 & 69.33 \\
 & VABench-point             & 54.71 & 59.33 & 4.67 & 48.17 & \textbf{72.34} & 27.56 & \underline{60.59} \\
 & PixMoPointsEval$^{\dagger}$ & 62.42 & \textbf{74.06} & 63.29 & 51.91 & \underline{65.96} & 61.25 & 57.94 \\
 & where2place               & 57.61 & \underline{73.52} & 66.51 & 57.86 & 73.15 & \textbf{74.63} & 60.74 \\
 & ShareRobot-Affor.     & \underline{29.22} & \textbf{30.84} & 27.58 & 26.26 & 19.39 & 22.61 & 26.19 \\
 & PIO                       & \underline{64.40} & \textbf{71.60} & 58.13 & 54.43 & 62.80 & 64.33 & 62.17 \\
 & NaviTrace$^{\ddagger}$    & \underline{30.87} & \textbf{42.80} & -25.27 & -11.18 & -25.77 & -12.39 & -36.42 \\
 & VABench-trace$^{\S}$ $\downarrow$ & 139.28 & \underline{108.15} & 173.44 & 173.29 & \textbf{85.57} & 177.42 & 132.40 \\
 & ShareRobot-Traj.$^{\S}$ $\downarrow$ & 0.3518 & \underline{0.2390} & 0.3061 & 0.4169 & 0.3052 & 0.2757 & \textbf{0.2317} \\
\cmidrule(lr){1-9}
\multirow{5}{=}{\raggedright\bfseries\textcolor{XPNavy}{Temporal Understanding}}
 & EgoTempo                  & \underline{38.20} & \textbf{44.60} & 30.00 & 22.40 & 32.60 & 32.40 & 19.20 \\
 & Video-MME                 & \underline{71.33} & \textbf{74.19} & 66.67 & 65.78 & 65.44 & 68.30 & 62.81 \\
 & MVBench                   & \underline{68.55} & \textbf{71.32} & 68.53 & 58.13 & 61.87 & 67.89 & 63.29 \\
 & LongVideoBench            & 60.48 & \textbf{64.89} & \underline{64.14} & 57.49 & 59.48 & 61.23 & 59.40 \\
 & QVHighlights-TimeLens     & \underline{54.82} & \textbf{58.26} & 39.37 & 4.55 & 47.53 & 31.22 & 35.07 \\
\cmidrule(lr){1-9}
\multirow{2}{=}{\raggedright\bfseries\textcolor{XPNavy}{State Verification}}
 & StateBench-P              & 65.80 & \textbf{76.80} & \underline{73.40} & 61.40 & 60.60 & 61.20 & 55.80 \\
 & StateBench-T              & \underline{43.76} & \textbf{46.21} & 33.39 & 34.39 & 32.06 & 37.52 & 37.38 \\
\cmidrule(lr){1-9}
\multirow{8}{=}{\raggedright\bfseries\textcolor{XPNavy}{General}}
 & MMMU                      & \underline{74.86} & \textbf{76.19} & 57.62 & 63.05 & 54.95 & 58.86 & 69.14 \\
 & MMVet                     & \textbf{80.09} & \underline{79.45} & 58.07 & 72.11 & 65.28 & 61.42 & 69.40 \\
 & RealWorldQA               & \underline{82.09} & \textbf{83.40} & 73.46 & 68.50 & 69.80 & 76.73 & 76.99 \\
 & MMBench-EN                & \underline{92.35} & \textbf{92.40} & 90.30 & 89.03 & 88.89 & 91.15 & 92.26 \\
 & IFEval                    & \textbf{92.42} & \underline{91.31} & 69.87 & 76.52 & 75.42 & 87.43 & 80.04 \\
 & MMLU-Pro                  & \underline{84.39} & \textbf{84.77} & 60.06 & 58.36 & 59.98 & 79.23 & 77.19 \\
 & BFCL-v3                   & \underline{58.75} & \textbf{60.12} & 10.00 & 6.88 & 23.75 & 46.88 & - \\
 & LiveCodeBench v6          & \textbf{77.31} & \underline{72.91} & 30.87 & 37.00 & 28.63 & 47.80 & 44.27 \\
\bottomrule
\end{tabularx}
\endgroup
\vspace{3pt}
\begin{minipage}{\textwidth}
\scriptsize\color{black!75}%
All results are reproduced under the DeepInsight evaluation protocol; the BFCL entry for Hy-Embodied-VLM-1.0 is left blank because its native tool-calling format is not currently supported. ShareRobot-Affor. and ShareRobot-Traj. denote ShareRobot-Affordance and ShareRobot-Trajectory, respectively. Unless otherwise specified, all entries report the native benchmark metric (higher is better, in \%). Exceptions include: $^{\dagger}$ point-level F1 (\%, higher is better); $^{\ddagger}$ NaviTrace official semantic-aware normalized score (higher is better and may be negative); and $^{\S}$ distance-based metrics, including VABench-trace trajectory RMSE (pixels) and ShareRobot-Trajectory discrete Fr\'echet distance (DFD), where lower is better. Bold and underlined values denote the best and second-best results. The blue column highlights \xvista.
\end{minipage}
\end{table}

Tables~\ref{tab:public_results_large} and~\ref{tab:public_results_compact} evaluate whether the execution-centric taxonomy leads to measurable gains in the capabilities repeatedly invoked during embodied execution. We treat the 35B-A3B track in Table~\ref{tab:public_results_large} as the primary comparison and use the 2B track in Table~\ref{tab:public_results_compact} to check whether the same training recipe yields a similar profile at smaller capacity. Within each table, the protocol-matched \xvista--Qwen pair is the main evidence for post-training effects; external models provide scale-aware context. Because the rows preserve native task formats, including accuracy, mean IoU, point F1, normalized trajectory score, RMSE, and DFD, we analyze directional row-wise changes rather than averaging heterogeneous metrics.

On the primary 35B-A3B track, \xvista improves 28 of 34 matched rows over the Qwen3.6 initialization. The gains concentrate on the execution-facing capabilities targeted by \xvista. In Action Guidance, all 10 rows improve, covering referring and point grounding (PixMoPointsEval, $62.42\!\rightarrow\!74.06$), affordance placement (Where2Place, $57.61\!\rightarrow\!73.52$), ordered navigation traces (NaviTrace, $30.87\!\rightarrow\!42.80$), and trajectory error metrics (VABench-trace RMSE decreases by 31.13 pixels; ShareRobot-Trajectory DFD decreases by 0.1128). These results show that the Guidance specialist improves not only recognition-style grounding, but also action-oriented targets and paths that can be consumed by downstream embodied systems.

The other capability families support the same execution-centric view. Temporal Understanding improves on all five evaluated rows, indicating stronger event and long-video reasoning. Both State Verification rows improve as well, with Capek-StateBench-P increasing by 11.00 points and Capek-StateBench-T by 2.45 points, showing that the model better evaluates physical predicates and task progress after post-training. Spatial Reasoning improves on 6 of 9 rows, including embodied and relation-heavy benchmarks such as VSI-Bench, MindCube, RoboSpatial-Home, EmbSpatial-Bench, and OpenEQA. General retention remains largely preserved, with 5 of 8 controls improving. Taken together, the 35B-A3B results indicate that \xvista strengthens the perception--reasoning--guidance--verification loop central to embodied execution while maintaining broad general competence.

\begin{table}[t]
\centering
\caption{Benchmark results for the 2B variant, compared with models of at most 4B parameters.}
\label{tab:public_results_compact}
\begingroup
\scriptsize
\setlength{\tabcolsep}{3.2pt}
\renewcommand{\arraystretch}{1.03}
\begin{tabularx}{0.99\textwidth}{@{}>{\raggedright\arraybackslash}p{0.145\linewidth}>{\raggedright\arraybackslash}p{0.235\linewidth}>{\centering\arraybackslash}X>{\columncolor{XPLink!12}\centering\arraybackslash}X>{\centering\arraybackslash}X>{\centering\arraybackslash}X@{}}
\toprule
\textbf{Task family} & \textbf{Benchmark} &
\shortstack{Qwen3.5\\2B} &
\shortstack{\textcolor{XPNavy}{\textbf{Capek0.5}}\\\textcolor{XPNavy}{\textbf{2B}}} &
\shortstack{RynnBrain1.1\\2B} &
\shortstack{RoboBrain\\2.5-4B} \\
\midrule
\multirow{10}{=}{\raggedright\bfseries\textcolor{XPNavy}{Spatial Reasoning}}
 & CV-Bench                 & 83.10 & 86.02 & \underline{86.49} & \textbf{86.90} \\
 & VSI-Bench                & 37.05 & \underline{55.00} & \textbf{73.83} & 49.80 \\
 & OmniSpatial              & \underline{46.18} & \textbf{50.16} & 39.73 & 41.62 \\
 & MindCube                 & 36.00 & \underline{47.62} & \textbf{54.86} & 30.48 \\
 & RoboSpatial-Home         & 25.52 & 38.13 & \textbf{59.95} & \underline{39.71} \\
 & All-Angles-Bench         & \underline{49.11} & \textbf{53.38} & 44.28 & 47.37 \\
 & EmbSpatial-Bench         & \underline{76.13} & \textbf{78.68} & 72.80 & 73.30 \\
 & ERQA                     & 35.00 & \textbf{46.25} & 41.25 & \underline{43.75} \\
 & OpenEQA                  & 51.48 & \textbf{57.78} & 50.70 & \underline{56.07} \\
\cmidrule(lr){1-6}
\multirow{10}{=}{\raggedright\bfseries\textcolor{XPNavy}{Action Guidance}}
 & RoboRefIt                & 76.50 & \textbf{79.89} & \underline{78.97} & 3.01 \\
 & PointBench               & 56.83 & \textbf{68.55} & 51.76 & \underline{67.13} \\
 & VABench-point            & 7.13 & \textbf{36.00} & 14.44 & \underline{25.27} \\
 & PixMoPointsEval$^{\dagger}$ & 33.90 & \underline{52.94} & 52.01 & \textbf{61.02} \\
 & where2place              & 29.16 & 35.86 & \underline{66.46} & \textbf{70.42} \\
 & ShareRobot-Affordance    & 18.33 & \underline{29.18} & 24.33 & \textbf{50.43} \\
 & PIO                      & 25.17 & 49.76 & \textbf{59.60} & \underline{58.43} \\
 & NaviTrace$^{\ddagger}$   & -42.08 & \textbf{12.59} & -20.74 & \underline{-20.43} \\
 & VABench-trace$^{\S}$ $\downarrow$  & 255.47 & \underline{160.63} & 216.83 & \textbf{155.10} \\
 & ShareRobot-Trajectory$^{\S}$ $\downarrow$ & 0.5689 & \underline{0.2683} & 0.2823 & \textbf{0.1619} \\
\cmidrule(lr){1-6}
\multirow{5}{=}{\raggedright\bfseries\textcolor{XPNavy}{Temporal Understanding}}
 & EgoTempo                 & 29.60 & \underline{33.40} & 23.40 & \textbf{34.40} \\
 & Video-MME                & 50.70 & \underline{61.30} & 59.96 & \textbf{62.78} \\
 & MVBench                  & 47.79 & 58.58 & \textbf{60.47} & \underline{60.08} \\
 & LongVideoBench           & 48.00 & \underline{58.07} & 54.33 & \textbf{58.65} \\
 & QVHighlights-TimeLens    & \underline{46.62} & 44.72 & 28.09 & \textbf{51.72} \\
\cmidrule(lr){1-6}
\multirow{2}{=}{\raggedright\bfseries\textcolor{XPNavy}{State Verification}}
 & StateBench-P  & 56.00 & \textbf{65.40} & 59.00 & \underline{59.20} \\
 & StateBench-T      & 25.52 & \textbf{33.46} & \underline{27.92} & 26.79 \\
\cmidrule(lr){1-6}
\multirow{7}{=}{\raggedright\bfseries\textcolor{XPNavy}{General}}
 & MMMU                     & 56.86 & \textbf{59.71} & 44.29 & \underline{53.05} \\
 & MMVet                    & 62.02 & \underline{62.48} & 46.51 & \textbf{65.97} \\
 & RealWorldQA              & \textbf{76.08} & \underline{75.69} & 72.03 & 68.76 \\
 & MMBench-EN               & \underline{88.45} & 88.20 & 86.90 & \textbf{88.54} \\
 & IFEval                   & \underline{81.52} & \textbf{81.70} & 63.59 & 70.61 \\
 & MMLU-Pro                 & \textbf{62.57} & \underline{61.30} & 37.57 & 54.20 \\
 & BFCL-v3                & \underline{25.37} & \textbf{25.75} & 2.00 & 0.13 \\
 & LiveCodeBench v6       & 11.01 & \textbf{26.65} & 10.79 & \underline{16.96} \\
\bottomrule
\end{tabularx}
\endgroup
\vspace{3pt}
\begin{minipage}{\textwidth}
\scriptsize\color{black!75}%
All results are reproduced under the DeepInsight evaluation protocol. Unless otherwise specified, all entries report the native benchmark metric (higher is better, in \%). Exceptions include: $^{\dagger}$ point-level F1 (\%, higher is better); $^{\ddagger}$ NaviTrace official semantic-aware normalized score (higher is better and may be negative); and $^{\S}$ distance-based metrics, including VABench-trace trajectory RMSE (pixels) and ShareRobot-Trajectory discrete Fr\'echet distance (DFD), where lower is better. Bold and underlined values denote the best and second-best results. The blue column highlights \xvista.
\end{minipage}
\end{table}

The compact 2B track shows a similar pattern under a different initialization and capacity regime. \xvistatwo improves 30 of 34 matched rows, including all 9 Spatial Reasoning rows, all 10 Action Guidance rows, 4 of 5 Temporal Understanding rows, both State Verification rows, and 5 of 8 general-retention controls. The largest gains again appear on execution-facing outputs: VABench-point increases by 28.87 points, PIO by 24.59 points, NaviTrace by 54.67 points, VABench-trace RMSE decreases by 94.84 pixels, and ShareRobot-Trajectory DFD decreases by 0.3006. This capacity-separated result suggests that the proposed capability organization is not specific to the 35B-A3B backbone, but transfers to a smaller dense model as the same execution-oriented pattern of improvement.

Overall, the broad benchmark results show that execution-centric post-training across the key stages of embodied execution yields a unified embodied model with stronger spatial, temporal, guidance, and state-verification capabilities, while largely preserving general multimodal and language abilities. These gains strengthen the perception--reasoning--guidance--verification loop required by embodied agents, providing a capability basis for the end-to-end task-success improvements examined in the following Embodied Agent Evaluation.

\subsection{Consolidation Analysis}
\label{sec:expert_integration}


Table~\ref{tab:expert_integration} presents a controlled consolidation ablation on the 35B-A3B track, where each capability domain is analyzed using representative signature benchmarks. All columns are initialized from the same Qwen3.6-35B-A3B backbone and are evaluated with the matched prompt, preprocessing, parser, and scorer for each row, so differences reflect how capability-specific behaviors are acquired and retained during integration. The shared start gives the pre-specialization reference, the matching expert provides the capability-acquisition reference for each row, and Mix-RL, TIES, MOPD, and TIES+MOPD compare alternative routes for consolidating the specialists into a single model.

\begin{table}[t]
\centering
\scriptsize
\setlength{\tabcolsep}{2.8pt}
\renewcommand{\arraystretch}{1.12}
\caption{
Capability specialization and consolidation on capability-specific 35B-A3B benchmark suites. All model variants are initialized from the same Qwen3.6-35B-A3B checkpoint and evaluated under a unified protocol with identical prompts, preprocessing, parsers, and scoring. Mix-RL denotes the data-mixing reinforcement learning baseline. All entries report the native benchmark metric; higher values indicate better performance except for VABench-trace, which reports trajectory RMSE (lower is better).
}
\label{tab:expert_integration}
\begin{tabularx}{0.98\linewidth}{
    @{}
    >{\raggedright\arraybackslash}p{0.12\linewidth}
    >{\raggedright\arraybackslash}X
    >{\centering\arraybackslash}p{0.098\linewidth}
    >{\centering\arraybackslash}p{0.098\linewidth}
    >{\centering\arraybackslash}p{0.085\linewidth}
    >{\centering\arraybackslash}p{0.085\linewidth}
    >{\centering\arraybackslash}p{0.085\linewidth}
    >{\columncolor{XPLink!10}\centering\arraybackslash}p{0.115\linewidth}
    @{}
}
\toprule
& &
\multicolumn{2}{c}{\textbf{Specialization}} &
\multicolumn{4}{c}{\textbf{Consolidation}} \\
\cmidrule(lr){3-4}
\cmidrule(lr){5-8}
\textbf{Capability} & \textbf{Signature benchmark} &
\shortstack{\textbf{Shared}\\\textbf{start}} &
\shortstack{\textbf{Matching}\\\textbf{expert}} &
\textbf{Mix-RL} & \textbf{TIES} & \textbf{MOPD} &
\shortstack{\textcolor{XPNavy}{\textbf{TIES}}\\[-1pt]\textcolor{XPNavy}{\textbf{+MOPD}}} \\
\midrule
\multirow{2}{*}{Spatial}
& MindCube & 58.67 & \textbf{71.52} & 60.38 & 58.38 & 68.48 & \underline{69.90} \\
& VSI-Bench & 60.83 & \underline{70.99} & 65.05 & \textbf{71.02} & 69.54 & 70.69 \\
\cmidrule(lr){1-8}
\multirow{2}{*}{Temporal}
& EgoTempo & 38.20 & \textbf{50.80} & 42.80 & \underline{49.40} & 43.20 & 44.60 \\
& LongVideo\allowbreak Bench & 60.48 & \textbf{65.31} & 63.31 & \underline{65.14} & 64.48 & 64.89 \\
\cmidrule(lr){1-8}
\multirow{2}{*}{Guidance}
& VABench-point & 54.71 & 59.00 & 58.33 & 57.67 & \textbf{59.78} & \underline{59.33} \\
& VABench-trace $\downarrow$ & 139.28 & \textbf{100.43} & 121.8 & 117.49 & 108.83 & \underline{108.15} \\
\cmidrule(lr){1-8}
\multirow{2}{*}{State}
& StateBench-\allowbreak P & 65.80 & \textbf{80.00} & 70.40 & 68.00 & 75.80 & \underline{76.80} \\
& StateBench-\allowbreak T & 43.76 & \underline{47.88} & \textbf{48.23} & 43.55 & 45.52 & 46.21 \\
\bottomrule
\end{tabularx}
\vspace{2pt}
\end{table}

The results in Table~\ref{tab:expert_integration} lead to three main observations. First, domain-specific post-training produces substantial and consistent specialization gains. Starting from the same shared backbone, the matching experts improve all eight signature benchmarks, including gains of (+12.85) on MindCube, (+10.16) on VSI-Bench, (+12.60) on EgoTempo, and (+14.20) on StateBench-P, together with a (38.85)-point reduction in VABench-trace RMSE. These specialized experts then serve as the source checkpoints for TIES merging and as the teacher policies for MOPD, while their improvements over the shared start define the capability gains that the consolidation stage aims to preserve. Second, the individual consolidation routes exhibit different capability biases. Mix-RL improves over the shared start across all benchmarks but achieves the best result only on StateBench-T, suggesting broad yet relatively limited retention of the expert gains. TIES preserves spatial and temporal capabilities particularly well, achieving the strongest consolidated results on VSI-Bench, EgoTempo, and LongVideoBench, but introduces regressions on MindCube and StateBench-T relative to the shared start. MOPD provides a more uniform profile and performs especially well on action- and state-oriented tasks, including the best VABench-point result and strong performance on VABench-trace and StateBench-P, although it retains less of the temporal expert gains than TIES. Third, applying MOPD from the TIES-merged initialization yields the strongest overall consolidation trade-off. TIES+MOPD achieves the best consolidated results on MindCube, VABench-trace, and StateBench-P, and ranks second on each of the remaining five benchmarks. Relative to MOPD alone, it improves seven of the eight results while giving up only (0.45) points on VABench-point. Taken together, these results favor TIES+MOPD as the final consolidation strategy: starting from the TIES-merged model allows subsequent MOPD training to retain consistently strong performance across all four capability domains while mitigating the capability trade-offs observed with either consolidation route alone.

\subsection{Embodied Agent Evaluation}
\label{sec:vigil_evaluation}

Beyond offline reasoning benchmarks, we evaluate whether the capabilities learned by \xvista transfer to embodied task execution. We therefore evaluate the 35B-A3B model variant in two complementary simulated environments: EmbodiedBench~\citep{yang2025embodiedbench}, which measures end-to-end embodied task success, and \vigil~\citep{vigil2026}, which provides fine-grained diagnosis of both isolated capabilities and their composition during interaction. Both benchmarks are strictly held out for evaluation, and no trajectories, observations, actions, or environment feedback are used during expert training or consolidation.

\smallskip\noindent\textbf{EmbodiedBench.}
We evaluate \xvista on the two high-level suites of EmbodiedBench, EB-Habitat (EB-HAB) and EB-ALFRED (EB-ALF), each comprising 300 multi-step embodied episodes. Performance is measured by task success rate (SR).
All models are evaluated using the same episode sets and default benchmark settings.

\begin{table}[t]
\centering
\scriptsize
\setlength{\tabcolsep}{3.2pt}
\renewcommand{\arraystretch}{1.12}
\caption{EmbodiedBench evaluation on EB-HAB and EB-ALF.
All values are task success rates (\%). Bold and underlined values
denote the best and second-best results, respectively.}
\label{tab:embodiedbench}
\resizebox{0.99\linewidth}{!}{
\begin{tabular}{@{}l*{14}{c}@{}}
\toprule
& \multicolumn{7}{c}{\textbf{EB-HAB}}
& \multicolumn{7}{c}{\textbf{EB-ALF}} \\
\cmidrule(lr){2-8}
\cmidrule(lr){9-15}
\textbf{Model}
& \textbf{Avg.}
& \textbf{Base}
& \textbf{Comm.}
& \textbf{Comp.}
& \textbf{Spatial}
& \textbf{Visual}
& \textbf{Long}
& \textbf{Avg.}
& \textbf{Base}
& \textbf{Comm.}
& \textbf{Comp.}
& \textbf{Spatial}
& \textbf{Visual}
& \textbf{Long} \\
\midrule

Embodied-R1.5
& 21.7
& 56.0
& 6.0
& 18.0
& 22.0
& 12.0
& 16.0
& 14.7
& 20.0
& 18.0
& 14.0
& 18.0
& 12.0
& 6.0 \\

RynnBrain1.1-9B
& 44.7
& 82.0
& 30.0
& 52.0
& \underline{34.0}
& \underline{48.0}
& 22.0
& 35.3
& 46.0
& 42.0
& 38.0
& 28.0
& 20.0
& 38.0 \\

RoboBrain2.0-32B
& 38.3
& 68.0
& 22.0
& \underline{54.0}
& 22.0
& 42.0
& 22.0
& 25.0
& 34.0
& 24.0
& 34.0
& 20.0
& 26.0
& 12.0 \\

RynnBrain-30B-A3B
& 26.7
& 66.0
& 14.0
& 34.0
& 22.0
& 22.0
& 2.0
& 26.3
& 32.0
& 30.0
& 34.0
& 22.0
& 26.0
& 14.0 \\

Qwen3.6-35B-A3B
& \underline{46.0}
& \underline{86.0}
& \underline{32.0}
& \underline{54.0}
& 32.0
& 44.0
& \underline{28.0}
& \underline{50.7}
& \textbf{52.0}
& \textbf{60.0}
& \underline{48.0}
& \textbf{44.0}
& \underline{44.0}
& \underline{56.0} \\

\rowcolor{XPLink!10}
\textbf{\xvistalarge}
& \textbf{63.0}
& \textbf{96.0}
& \textbf{48.0}
& \textbf{74.0}
& \textbf{46.0}
& \textbf{60.0}
& \textbf{54.0}
& \textbf{55.3}
& \textbf{52.0}
& \underline{58.0}
& \textbf{70.0}
& \underline{42.0}
& \textbf{48.0}
& \textbf{62.0} \\


\bottomrule
\end{tabular}
}
\vspace{2pt}

\begin{minipage}{0.99\linewidth}
\scriptsize\color{black!75}
Comm.: Common Sense; Comp.: Complex Instruction; Spatial: Spatial Awareness; Visual: Visual Appearance; Long:
Long Horizon. 
Each subset contains 50 episodes. 
RynnBrain 1.1, Qwen 3.6, and \xvista are evaluated with thinking mode enabled.
Embodied-R1.5, RoboBrain 2.0, and RynnBrain are evaluated using their respective native inference configurations.
\end{minipage}
\end{table}

As shown in Table~\ref{tab:embodiedbench}, \xvista achieves the highest average success rate on both EB-HAB and EB-ALF, outperforming the scale-matched Qwen3.6 backbone by 17.0 and 4.6 percentage points, respectively. The gains are especially large on long-horizon tasks, where success increases by 26.0 points on EB-HAB and 6.0 points on EB-ALF. \xvista also improves the complex-instruction subsets in both environments (+20.0 on EB-HAB and +22.0 on EB-ALF). These gains indicate that consolidating complementary reasoning capabilities improves the maintenance of task context and the coordination of perception, planning, and interaction over extended execution sequences.

\smallskip\noindent\textbf{VIGIL.}
We further evaluate \xvista on VIGIL, a frozen embodied-agent benchmark comprising 1,000 balanced episodes in AI2-THOR~\citep{kolve2017ai2thor} and ProcTHOR~\citep{deitke2022procthor}. VIGIL distinguishes successful interaction with the environment from correct terminal reporting by separately evaluating world completion and benchmark success.
Following the original protocol, we report overall world completion ($W$), overall benchmark success ($B$), and benchmark success for each task family. Here, $W$ measures whether the target world predicate is satisfied, whereas $B$ additionally requires a correct terminal report. The four diagnostic families evaluate grounding, approach, search, and state verification in isolation, while the four compositional families assess how these capabilities combine through interaction, multi-step manipulation, and constraint resolution.

\begin{table}[t]
\centering
\scriptsize
\setlength{\tabcolsep}{1.15pt}
\renewcommand{\arraystretch}{1.14}
\caption{VIGIL agentic evaluation. Overall columns report primary score (Score), world completion ($W$), and benchmark success ($B$); the eight task-family columns report $W$/$B$. All values are percentages.}
\label{tab:vigil_downstream_wb}
\begin{tabularx}{0.99\linewidth}{@{}>{\raggedright\arraybackslash}X*{3}{>{\centering\arraybackslash}p{0.038\linewidth}}*{8}{>{\centering\arraybackslash}p{0.071\linewidth}}@{}}
\toprule
& \multicolumn{3}{c}{\textbf{Overall}} & \multicolumn{4}{c}{\textbf{Diagnostic Probes $W$/$B$}} & \multicolumn{4}{c}{\textbf{Compositional Tasks $W$/$B$}} \\
\cmidrule(lr){2-4}\cmidrule(lr){5-8}\cmidrule(lr){9-12}
\textbf{Model} & \textbf{Score} & \textbf{$W$} & \textbf{$B$} & \textbf{PG} & \textbf{DA} & \textbf{VS} & \textbf{SV} & \textbf{AI} & \textbf{SI} & \textbf{SM} & \textbf{CR} \\
\midrule
Embodied-R1.5 & 31.7 & 33.4 & 31.7 & 74.4/\textbf{74.4} & \textbf{59.2}/\textbf{58.4} & 10.4/10.4 & \textbf{96.0}/83.2 & 14.4/14.4 & 0.8/0.8 & \underline{6.4}/\underline{6.4} & 5.6/\underline{5.6} \\
Qwen3.6-35B-A3B & 30.2 & \underline{37.4} & 28.8 & \underline{76.0}/52.8 & 22.4/20.8 & 36.0/\underline{35.2} & \textbf{96.0}/\textbf{85.6} & \underline{41.6}/\underline{18.4} & \textbf{11.2}/\textbf{7.2} & 4.0/3.2 & \textbf{12.0}/\textbf{7.2} \\
RynnBrain1.1-9B & 29.1 & 34.3 & 23.4 & 73.6/48.0 & 36.0/26.4 & \underline{47.2}/16.8 & \textbf{96.0}/84.0 & 16.8/9.6 & 0.0/0.0 & 1.6/0.0 & 3.2/2.4 \\
RoboBrain2.0-32B & \textbf{38.1} & 36.2 & \textbf{34.3} & 71.2/69.6 & 32.8/32.8 & \textbf{71.2}/\textbf{71.2} & 90.4/76.8 & 12.0/12.0 & 2.4/2.4 & \textbf{8.0}/\textbf{8.0} & 1.6/1.6 \\
RynnBrain-30B-A3B & 17.2 & 22.2 & 13.5 & 6.4/5.6 & \underline{53.6}/\underline{41.6} & 19.2/10.4 & \underline{92.0}/46.4 & 3.2/1.6 & 0.0/0.0 & 3.2/2.4 & 0.0/0.0 \\
\rowcolor{XPLink!10}\textbf{\xvistalarge} & \underline{36.4} & \textbf{38.6} & \underline{32.2} & \textbf{85.6}/\underline{72.8} & 19.2/19.2 & 36.8/\underline{35.2} & \textbf{96.0}/\underline{84.8} & \textbf{47.2}/\textbf{28.8} & \underline{9.6}/\underline{4.0} & \underline{6.4}/5.6 & \underline{8.0}/\textbf{7.2} \\
\bottomrule
\end{tabularx}
\vspace{2pt}
\begin{minipage}{0.99\linewidth}
\scriptsize\color{black!75}Each family contains 125 episodes under the same episode pack, prompt, parser, observation-history policy, decoding configuration, and action budget. PG: pixel grounding; DA: distance approach; VS: view search; SV: state verification; AI: approach-and-interact; SI: search-and-interact; SM: sequential manipulation; CR: constraint resolving.
\end{minipage}
\end{table}


As shown in Table~\ref{tab:vigil_downstream_wb}, \xvista improves both world completion and benchmark success over the Qwen3.6 backbone, increasing overall $W$ from 37.4\% to 38.6\% and overall $B$ from 28.8\% to 32.2\%. The largest gain appears in pixel grounding, where PG increases from 76.0/52.8 to 85.6/72.8 in $W$/$B$, indicating more accurate localization of task-relevant visual targets. The gains extend to compositional tasks: benchmark success rises from 18.4\% to 28.8\% on approach-and-interact and from 3.2\% to 5.6\% on sequential manipulation. Since these tasks require the agent to identify relevant targets, approach them, execute actions, and verify the resulting state across multiple steps, the results suggest that execution-centric capability consolidation transfers beyond isolated reasoning benchmarks to embodied task execution.

Overall, the gains on both EmbodiedBench and VIGIL indicate that the execution-centric capabilities learned by \xvista compose effectively during embodied interaction, leading to more robust task-level execution.

\FloatBarrier

\FloatBarrier
\section{Conclusion and Future Work}
\label{sec:conclusion}

We presented \xvista, a unified vision-language model for execution-centric embodied reasoning. Its post-training recipe derives Spatial Reasoning, Temporal Understanding, Action Guidance, and State Verification experts from a shared checkpoint, matches each expert's training to its native supervision signals and output contracts, and consolidates their behaviors into a single inference-time model through weight-space merging followed by routed policy-space distillation. This design uses the recurring evidence needs of the execution cycle as an organizing principle, without imposing a rigid stage decomposition or treating the VLM as a low-level controller.

We evaluated this design at three complementary levels: public benchmark comparisons for the 2B and 35B-A3B tracks, a controlled study of capability retention from specialist experts to the unified model, and closed-loop evaluation in simulated embodied environments. Together, these evaluations connect broad embodied capability, specialist-to-unified retention, and capability composition during interactive task execution.

Looking forward, we aim to extend \xvista from a unified embodied reasoner into an agentic embodied brain that perceives, orchestrates, and acts through tools: perception tools for acquiring task-relevant evidence, executable code as a programmatic orchestration layer for long-horizon decomposition and recovery, and action tools that expose robot skills while leaving low-level control to specialized policies. In this setting, tool invocation becomes a first-class action within the execution cycle, and the model must decide when and which tools to call, compose calls into executable workflows, interpret feedback, and replan when outcomes deviate from expectations. We plan to acquire these abilities through new specialists and consolidate them under the same traceable specialist-to-unified protocol established in this report, with evaluation extended to longer-horizon interactive tasks and physical-robot settings.

\FloatBarrier

\clearpage
\bibliographystyle{unsrtnat}
\bibliography{refs}

\begin{thebibliography}{96}
\providecommand{\natexlab}[1]{#1}
\providecommand{\url}[1]{\texttt{#1}}
\expandafter\ifx\csname urlstyle\endcsname\relax
  \providecommand{\doi}[1]{doi: #1}\else
  \providecommand{\doi}{doi: \begingroup \urlstyle{rm}\Url}\fi

\bibitem[Radford et~al.(2021)Radford, Kim, Hallacy, Ramesh, Goh, Agarwal, Sastry, Askell, Mishkin, Clark, Krueger, and Sutskever]{radford2021clip}
Alec Radford, Jong~Wook Kim, Chris Hallacy, Aditya Ramesh, Gabriel Goh, Sandhini Agarwal, Girish Sastry, Amanda Askell, Pamela Mishkin, Jack Clark, Gretchen Krueger, and Ilya Sutskever.
\newblock Learning transferable visual models from natural language supervision.
\newblock In \emph{International Conference on Machine Learning (ICML)}, 2021.

\bibitem[Alayrac et~al.(2022)Alayrac, Donahue, Luc, et~al.]{alayrac2022flamingo}
Jean-Baptiste Alayrac, Jeff Donahue, Pauline Luc, et~al.
\newblock Flamingo: a visual language model for few-shot learning.
\newblock \emph{Advances in Neural Information Processing Systems (NeurIPS)}, 2022.

\bibitem[Li et~al.(2023{\natexlab{a}})Li, Li, Savarese, and Hoi]{li2023blip2}
Junnan Li, Dongxu Li, Silvio Savarese, and Steven Hoi.
\newblock Blip-2: Bootstrapping language-image pre-training with frozen image encoders and large language models.
\newblock In \emph{International Conference on Machine Learning (ICML)}, 2023{\natexlab{a}}.

\bibitem[Dai et~al.(2023)Dai, Li, Li, Tiong, Zhao, Wang, Li, Fung, and Hoi]{dai2023instructblip}
Wenliang Dai, Junnan Li, Dongxu Li, Anthony Meng~Huat Tiong, Junqi Zhao, Weisheng Wang, Boyang Li, Pascale Fung, and Steven Hoi.
\newblock Instructblip: Towards general-purpose vision-language models with instruction tuning.
\newblock In \emph{Advances in Neural Information Processing Systems (NeurIPS)}, 2023.

\bibitem[Liu et~al.(2023{\natexlab{a}})Liu, Li, Wu, and Lee]{liu2023llava}
Haotian Liu, Chunyuan Li, Qingyang Wu, and Yong~Jae Lee.
\newblock Visual instruction tuning.
\newblock In \emph{Advances in Neural Information Processing Systems (NeurIPS)}, 2023{\natexlab{a}}.

\bibitem[Wang et~al.(2024{\natexlab{a}})Wang, Bai, Tan, et~al.]{wang2024qwen2vl}
Peng Wang, Shuai Bai, Sinan Tan, et~al.
\newblock Qwen2-vl: Enhancing vision-language model's perception of the world at any resolution.
\newblock \emph{arXiv preprint arXiv:2409.12191}, 2024{\natexlab{a}}.

\bibitem[Bai et~al.(2025{\natexlab{a}})Bai, Chen, Liu, et~al.]{bai2025qwen25vl}
Shuai Bai, Keqin Chen, Xuejing Liu, et~al.
\newblock Qwen2.5-vl technical report.
\newblock \emph{arXiv preprint arXiv:2502.13923}, 2025{\natexlab{a}}.

\bibitem[Ahn et~al.(2022)Ahn, Brohan, Brown, et~al.]{saycan2022}
Michael Ahn, Anthony Brohan, Noah Brown, et~al.
\newblock Do as i can, not as i say: Grounding language in robotic affordances.
\newblock \emph{arXiv preprint arXiv:2204.01691}, 2022.

\bibitem[Driess et~al.(2023)Driess, Xia, Sajjadi, et~al.]{palme2023}
Danny Driess, Fei Xia, Mehdi S.~M. Sajjadi, et~al.
\newblock Palm-e: An embodied multimodal language model.
\newblock \emph{arXiv preprint arXiv:2303.03378}, 2023.

\bibitem[Liang et~al.(2022)Liang, Huang, Xia, et~al.]{codeaspolicies2022}
Jacky Liang, Wenlong Huang, Fei Xia, et~al.
\newblock Code as policies: Language model programs for embodied control.
\newblock \emph{arXiv preprint arXiv:2209.07753}, 2022.

\bibitem[Huang et~al.(2023)Huang, Wang, Zhang, et~al.]{voxposer2023}
Wenlong Huang, Chen Wang, Ruohan Zhang, et~al.
\newblock Voxposer: Composable 3d value maps for robotic manipulation with language models.
\newblock \emph{arXiv preprint arXiv:2307.05973}, 2023.

\bibitem[{Gemini Robotics Team}(2025)]{geminirobotics2025}
{Gemini Robotics Team}.
\newblock Gemini robotics: Bringing ai into the physical world.
\newblock \emph{arXiv preprint arXiv:2503.20020}, 2025.

\bibitem[Yuan et~al.(2024)Yuan, Duan, Blukis, et~al.]{robopoint2024}
Wentao Yuan, Jiafei Duan, Valts Blukis, et~al.
\newblock Robopoint: A vision-language model for spatial affordance prediction for robotics.
\newblock \emph{arXiv preprint arXiv:2406.10721}, 2024.

\bibitem[Ji et~al.(2025)Ji, Tan, Shi, et~al.]{robobrain2025}
Yuheng Ji, Huajie Tan, Jiayu Shi, et~al.
\newblock Robobrain: A unified brain model for robotic manipulation from abstract to concrete.
\newblock \emph{arXiv preprint arXiv:2502.21257}, 2025.

\bibitem[Tan et~al.(2026)Tan, Zhou, Li, et~al.]{robobrain25_2026}
Huajie Tan, Enshen Zhou, Zhiyu Li, et~al.
\newblock Robobrain 2.5: Depth in sight, time in mind.
\newblock \emph{arXiv preprint arXiv:2601.14352}, 2026.

\bibitem[Dang et~al.(2026)Dang, Guo, Hou, et~al.]{rynnbrain2026}
Ronghao Dang, Jiayan Guo, Bohan Hou, et~al.
\newblock Rynnbrain: Open embodied foundation models.
\newblock \emph{arXiv preprint arXiv:2602.14979}, 2026.

\bibitem[Wang et~al.(2026{\natexlab{a}})Wang, Yu, Rao, Ling, Li, et~al.]{hyembodied10_2026}
Ziyi Wang, Xumin Yu, Yongming Rao, Yonggen Ling, Yunheng Li, et~al.
\newblock Hy-embodied-vlm-1.0: Efficient physical-world agents.
\newblock \emph{arXiv preprint arXiv:2607.12894}, 2026{\natexlab{a}}.

\bibitem[{NVIDIA}(2026)]{cosmos3_2026}
{NVIDIA}.
\newblock Cosmos 3: Omnimodal world models for physical ai.
\newblock \emph{arXiv preprint arXiv:2606.02800}, 2026.
\newblock URL \url{https://research.nvidia.com/labs/cosmos-lab/cosmos3/technical-report.pdf}.

\bibitem[Hao et~al.(2025{\natexlab{a}})Hao, Zhou, Huang, et~al.]{mimoembodied2025}
Xiaoshuai Hao, Lei Zhou, Zhijian Huang, et~al.
\newblock Mimo-embodied: X-embodied foundation model technical report.
\newblock \emph{arXiv preprint arXiv:2511.16518}, 2025{\natexlab{a}}.

\bibitem[Yuan et~al.(2026)Yuan, Huang, Yao, et~al.]{embodiedr15}
Yifu Yuan, Yaoting Huang, Xianze Yao, et~al.
\newblock Embodied-r1.5: Evolving physical intelligence via embodied foundation models.
\newblock \emph{arXiv preprint arXiv:2606.11324}, 2026.

\bibitem[Bjorck et~al.(2026)Bjorck, Li, Man, et~al.]{vesta2026}
Johan Bjorck, Zhiqi Li, Yunze Man, et~al.
\newblock Vesta: A generalist embodied reasoning model.
\newblock \emph{arXiv preprint arXiv:2606.20905}, 2026.

\bibitem[{ACE-Brain Team} et~al.(2026){ACE-Brain Team}, Gong, Gu, Luo, et~al.]{acebrain2026}
{ACE-Brain Team}, Ziyang Gong, Haoming Gu, Zehang Luo, et~al.
\newblock Ace-brain-0.5: A unified embodied foundational model for physical agentic ai.
\newblock \emph{arXiv preprint arXiv:2607.04426}, 2026.

\bibitem[Yang et~al.(2025{\natexlab{a}})Yang, Chen, Zhang, Zhao, Qian, Wang, Wang, Koripella, Movahedi, Li, et~al.]{yang2025embodiedbench}
Rui Yang, Hanyang Chen, Junyu Zhang, Mark Zhao, Cheng Qian, Kangrui Wang, Qineng Wang, Teja~Venkat Koripella, Marziyeh Movahedi, Manling Li, et~al.
\newblock Embodiedbench: Comprehensive benchmarking multi-modal large language models for vision-driven embodied agents.
\newblock \emph{arXiv preprint arXiv:2502.09560}, 2025{\natexlab{a}}.

\bibitem[Salimpour et~al.(2025)Salimpour, Fu, Rachwa{\l}, Bertrand, O'Sullivan, Jakob, Keramat, Militano, Toffetti, Edelman, et~al.]{salimpour2025towards}
Sahar Salimpour, Lei Fu, Kajetan Rachwa{\l}, Pascal Bertrand, Kevin O'Sullivan, Robert Jakob, Farhad Keramat, Leonardo Militano, Giovanni Toffetti, Harry Edelman, et~al.
\newblock Towards embodied agentic ai: Review and classification of llm-and vlm-driven robot autonomy and interaction.
\newblock \emph{arXiv preprint arXiv:2508.05294}, 2025.

\bibitem[Li et~al.(2023{\natexlab{b}})Li, Zhang, Wong, et~al.]{li2023behavior1k}
Chengshu Li, Ruohan Zhang, Josiah Wong, et~al.
\newblock Behavior-1k: A benchmark for embodied ai with 1,000 everyday activities and realistic simulation.
\newblock \emph{Conference on Robot Learning (CoRL)}, 2023{\natexlab{b}}.

\bibitem[Ma et~al.(2026)Ma, Wei, Zhao, et~al.]{mopd2026}
Wenhan Ma, Jianyu Wei, Liang Zhao, et~al.
\newblock Mopd: Multi-teacher on-policy distillation for capability integration in llm post-training.
\newblock \emph{arXiv preprint arXiv:2606.30406}, 2026.

\bibitem[Wang et~al.(2026{\natexlab{b}})Wang, Long, Li, Xu, Li, and Tang]{wang2026mix}
Haoqing Wang, Xiang Long, Ziheng Li, Yilong Xu, Tingguang Li, and Yehui Tang.
\newblock To mix or to merge: Toward multi-domain reinforcement learning for large language models.
\newblock \emph{arXiv preprint arXiv:2602.12566}, 2026{\natexlab{b}}.

\bibitem[Yadav et~al.(2023)Yadav, Tam, Choshen, et~al.]{ties2023}
Prateek Yadav, Derek Tam, Leshem Choshen, et~al.
\newblock Ties-merging: Resolving interference when merging models.
\newblock \emph{arXiv preprint arXiv:2306.01708}, 2023.

\bibitem[{Qwen Team}(2026{\natexlab{a}})]{qwen36_35b_a3b}
{Qwen Team}.
\newblock {Qwen3.6-35B-A3B}: Agentic coding power, now open to all, April 2026{\natexlab{a}}.
\newblock URL \url{https://qwen.ai/blog?id=qwen3.6-35b-a3b}.

\bibitem[{Qwen Team}(2026{\natexlab{b}})]{qwen3.5}
{Qwen Team}.
\newblock {Qwen3.5}: Towards native multimodal agents, February 2026{\natexlab{b}}.
\newblock URL \url{https://qwen.ai/blog?id=qwen3.5}.

\bibitem[Chen et~al.(2026)Chen, Fang, Jiang, Wang, Gu, Yi, and Chen]{vigil2026}
Ying Chen, Lihuang Fang, Rui Jiang, Mingxu Wang, Zhifeng Gu, Lei Yi, and Jie Chen.
\newblock Done, but not sure: Disentangling world completion from self-termination in embodied agents.
\newblock \emph{arXiv preprint arXiv:2605.08747}, 2026.

\bibitem[Liu et~al.(2025)Liu, Liang, Hu, Peng, Lu, Xu, Fu, Yin, et~al.]{liu2025spatial}
Disheng Liu, Tuo Liang, Zhe Hu, Jierui Peng, Yiren Lu, Yi~Xu, Yun Fu, Yu~Yin, et~al.
\newblock Spatial intelligence in vision-language models: A comprehensive survey.
\newblock 2025.

\bibitem[Cai et~al.(2026)Cai, Wang, Gu, Pu, Xu, Wang, Yin, Yang, Wei, Zhou, et~al.]{cai2026scaling}
Zhongang Cai, Ruisi Wang, Chenyang Gu, Fanyi Pu, Junxiang Xu, Yubo Wang, Wanqi Yin, Zhitao Yang, Chen Wei, Tongxi Zhou, et~al.
\newblock Scaling spatial intelligence with multimodal foundation models.
\newblock In \emph{Proceedings of the IEEE/CVF Conference on Computer Vision and Pattern Recognition}, pages 7879--7890, 2026.

\bibitem[Yang et~al.(2025{\natexlab{b}})Yang, Yang, Gupta, Han, Fei-Fei, and Xie]{yang2025thinkinginspace}
Jihan Yang, Shusheng Yang, Anjali~W. Gupta, Rilyn Han, Li~Fei-Fei, and Saining Xie.
\newblock Thinking in space: How multimodal large language models see, remember, and recall spaces.
\newblock In \emph{Proceedings of the IEEE/CVF Conference on Computer Vision and Pattern Recognition (CVPR)}, pages 10632--10643, June 2025{\natexlab{b}}.

\bibitem[Wang et~al.(2025)Wang, Yin, Zhang, et~al.]{wang2025mindcube}
Qineng Wang, Baiqiao Yin, Pingyue Zhang, et~al.
\newblock Mindcube: Spatial mental modeling from limited views.
\newblock \emph{arXiv preprint arXiv:2506.21458}, 2025.

\bibitem[Du et~al.(2024)Du, Wu, Li, Huang, and Wei]{du2024embspatial}
Mengfei Du, Binhao Wu, Zejun Li, Xuanjing Huang, and Zhongyu Wei.
\newblock Embspatial-bench: Benchmarking spatial understanding for embodied tasks with large vision-language models.
\newblock \emph{arXiv preprint arXiv:2406.05756}, 2024.

\bibitem[Yi et~al.(2019)Yi, Gan, Li, Kohli, Wu, Torralba, and Tenenbaum]{yi2019clevrer}
Kexin Yi, Chuang Gan, Yunzhu Li, Pushmeet Kohli, Jiajun Wu, Antonio Torralba, and Joshua~B Tenenbaum.
\newblock Clevrer: Collision events for video representation and reasoning.
\newblock \emph{arXiv preprint arXiv:1910.01442}, 2019.

\bibitem[Xiao et~al.(2021)Xiao, Shang, Yao, and Chua]{xiao2021nextqa}
Junbin Xiao, Xindi Shang, Angela Yao, and Tat-Seng Chua.
\newblock Next-qa: Next phase of question-answering to explaining temporal actions.
\newblock In \emph{Proceedings of the IEEE/CVF Conference on Computer Vision and Pattern Recognition (CVPR)}, 2021.

\bibitem[Patraucean et~al.(2023)Patraucean, Smaira, Gupta, Recasens, Markeeva, Banarse, Koppula, Malinowski, Yang, Doersch, et~al.]{patraucean2023perception}
Viorica Patraucean, Lucas Smaira, Ankush Gupta, Adria Recasens, Larisa Markeeva, Dylan Banarse, Skanda Koppula, Mateusz Malinowski, Yi~Yang, Carl Doersch, et~al.
\newblock Perception test: A diagnostic benchmark for multimodal video models.
\newblock \emph{Advances in Neural Information Processing Systems}, 36:\penalty0 42748--42761, 2023.

\bibitem[Wu et~al.(2024{\natexlab{a}})Wu, Yu, Chen, Tenenbaum, and Gan]{wu2024star}
Bo~Wu, Shoubin Yu, Zhenfang Chen, Joshua~B Tenenbaum, and Chuang Gan.
\newblock Star: A benchmark for situated reasoning in real-world videos.
\newblock \emph{arXiv preprint arXiv:2405.09711}, 2024{\natexlab{a}}.

\bibitem[Zhang et~al.(2024)Zhang, Wu, Li, et~al.]{zhang2024llavavideo}
Yuanhan Zhang, Jinming Wu, Wei Li, et~al.
\newblock Llava-video: Video instruction tuning with synthetic data.
\newblock \emph{arXiv preprint arXiv:2410.02713}, 2024.

\bibitem[Chen et~al.(2025{\natexlab{a}})Chen, Huang, Shi, Hu, Ye, Zhu, Liu, Molchanov, Kautz, Qi, Liu, Yin, Lu, and Han]{chen2025longvila-r1}
Yukang Chen, Wei Huang, Baifeng Shi, Qinghao Hu, Hanrong Ye, Ligeng Zhu, Zhijian Liu, Pavlo Molchanov, Jan Kautz, Xiaojuan Qi, Sifei Liu, Hongxu Yin, Yao Lu, and Song Han.
\newblock Scaling rl to long videos.
\newblock 2025{\natexlab{a}}.

\bibitem[Sigurdsson et~al.(2016)Sigurdsson, Varol, Wang, Farhadi, Laptev, and Gupta]{sigurdsson2016charades}
Gunnar~A. Sigurdsson, Gul Varol, Xiaolong Wang, Ali Farhadi, Ivan Laptev, and Abhinav Gupta.
\newblock Hollywood in homes: Crowdsourcing data collection for activity understanding.
\newblock In \emph{Proceedings of the European Conference on Computer Vision (ECCV)}, 2016.

\bibitem[Hendricks et~al.(2017)Hendricks, Wang, Shechtman, Sivic, Darrell, and Russell]{hendricks2017didemo}
Lisa~Anne Hendricks, Oliver Wang, Eli Shechtman, Josef Sivic, Trevor Darrell, and Bryan Russell.
\newblock Localizing moments in video with natural language.
\newblock In \emph{Proceedings of the IEEE International Conference on Computer Vision (ICCV)}, 2017.

\bibitem[Zala et~al.(2023)Zala, Cho, Kottur, Chen, Oguz, Mehdad, and Bansal]{zala2023hierarchical}
Abhay Zala, Jaemin Cho, Satwik Kottur, Xilun Chen, Barlas Oguz, Yashar Mehdad, and Mohit Bansal.
\newblock Hierarchical video-moment retrieval and step-captioning.
\newblock In \emph{Proceedings of the IEEE/CVF Conference on Computer Vision and Pattern Recognition}, pages 23056--23065, 2023.

\bibitem[Oncescu et~al.(2021)Oncescu, Henriques, Liu, Zisserman, and Albanie]{oncescu2021queryd}
Andreea-Maria Oncescu, Joao~F Henriques, Yang Liu, Andrew Zisserman, and Samuel Albanie.
\newblock Queryd: A video dataset with high-quality text and audio narrations.
\newblock In \emph{ICASSP 2021-2021 IEEE International Conference on Acoustics, Speech and Signal Processing (ICASSP)}, pages 2265--2269. IEEE, 2021.

\bibitem[Li et~al.(2025)Li, Chen, Wei, et~al.]{li2025llavast}
Hongyu Li, Jinyu Chen, Ziyu Wei, et~al.
\newblock Llava-st: A multimodal large language model for fine-grained spatial-temporal understanding.
\newblock \emph{arXiv preprint arXiv:2501.08282}, 2025.

\bibitem[Chen et~al.(2024)Chen, Wei, Li, et~al.]{chen2024sharegpt4video}
Lin Chen, Xilin Wei, Jinsong Li, et~al.
\newblock Sharegpt4video: Improving video understanding and generation with better captions.
\newblock \emph{arXiv preprint arXiv:2406.04325}, 2024.

\bibitem[Han et~al.(2025)Han, Huang, Shi, Zhuo, Su, Zhang, Zhou, Qi, Liao, and Liu]{han2025videoespresso}
Songhao Han, Wei Huang, Hairong Shi, Le~Zhuo, Xiu Su, Shifeng Zhang, Xu~Zhou, Xiaojuan Qi, Yue Liao, and Si~Liu.
\newblock Videoespresso: A large-scale chain-of-thought dataset for fine-grained video reasoning via core frame selection.
\newblock In \emph{Proceedings of the Computer Vision and Pattern Recognition Conference}, pages 26181--26191, 2025.

\bibitem[{NVIDIA} et~al.(2025){NVIDIA}, Azzolini, Bai, et~al.]{nvidia2025cosmosreason1}
{NVIDIA}, Alisson Azzolini, Junjie Bai, et~al.
\newblock Cosmos-reason1: From physical common sense to embodied reasoning.
\newblock \emph{arXiv preprint arXiv:2503.15558}, 2025.

\bibitem[Deitke et~al.(2024)Deitke, Clark, Lee, et~al.]{deitke2024molmo}
Matt Deitke, Christopher Clark, Sangho Lee, et~al.
\newblock Molmo and pixmo: Open weights and open data for state-of-the-art vision-language models.
\newblock \emph{arXiv preprint arXiv:2409.17146}, 2024.

\bibitem[Khazatsky et~al.(2024)Khazatsky, Pertsch, Nair, et~al.]{khazatsky2024droid}
Alexander Khazatsky, Karl Pertsch, Suraj Nair, et~al.
\newblock Droid: A large-scale in-the-wild robot manipulation dataset.
\newblock \emph{arXiv preprint arXiv:2403.12945}, 2024.

\bibitem[{AgiBot-World Contributors} et~al.(2025){AgiBot-World Contributors}, Bu, Cai, et~al.]{agibot2025colosseo}
{AgiBot-World Contributors}, Qingwen Bu, Jisong Cai, et~al.
\newblock Agibot world colosseo: A large-scale manipulation platform for scalable and intelligent embodied systems.
\newblock \emph{arXiv preprint arXiv:2503.06669}, 2025.

\bibitem[Hou et~al.(2025)Hou, Wu, Liu, et~al.]{hou2025robomind2}
Chengkai Hou, Kun Wu, Jiaming Liu, et~al.
\newblock Robomind 2.0: A multimodal, bimanual mobile manipulation dataset for generalizable embodied intelligence.
\newblock \emph{arXiv preprint arXiv:2512.24653}, 2025.

\bibitem[Yuan et~al.(2025{\natexlab{a}})Yuan, Cui, Chen, Dong, Ni, Kou, Liu, Li, Zheng, and Hao]{yuan2025fsd}
Yifu Yuan, Haiqin Cui, Yibin Chen, Zibin Dong, Fei Ni, Longxin Kou, Jinyi Liu, Pengyi Li, Yan Zheng, and Jianye Hao.
\newblock From seeing to doing: Bridging reasoning and decision for robotic manipulation.
\newblock \emph{arXiv preprint arXiv:2505.08548}, 2025{\natexlab{a}}.

\bibitem[Sermanet et~al.(2023)Sermanet, Ding, Zhao, et~al.]{sermanet2023robovqa}
Pierre Sermanet, Tianli Ding, Jeffrey Zhao, et~al.
\newblock Robovqa: Multimodal long-horizon reasoning for robotics.
\newblock \emph{arXiv preprint arXiv:2311.00899}, 2023.

\bibitem[Chen et~al.(2025{\natexlab{b}})Chen, Xie, Ma, Sanketi, and Goldberg]{chen2025robo2vlm}
Kaiyuan Chen, Shuangyu Xie, Zehan Ma, Pannag~R. Sanketi, and Ken Goldberg.
\newblock Robo2vlm: Visual question answering from large-scale in-the-wild robot manipulation datasets.
\newblock \emph{arXiv preprint arXiv:2505.15517}, 2025{\natexlab{b}}.

\bibitem[Hao et~al.(2025{\natexlab{b}})Hao, Tang, Zhang, et~al.]{hao2025roboafford}
Xiaoshuai Hao, Yingbo Tang, Lingfeng Zhang, et~al.
\newblock Roboafford++: A generative ai-enhanced dataset for multimodal affordance learning in robotic manipulation and navigation.
\newblock \emph{arXiv preprint arXiv:2511.12436}, 2025{\natexlab{b}}.

\bibitem[Zhang et~al.(2025{\natexlab{a}})Zhang, Chen, Zhou, Xu, Huang, Mei, Chen, Yuan, Cai, Huang, Quan, Xu, and Zhang]{zhang2025spar}
Jiahui Zhang, Yurui Chen, Yanpeng Zhou, Yueming Xu, Ze~Huang, Jilin Mei, Junhui Chen, Yujie Yuan, Xinyue Cai, Guowei Huang, Xingyue Quan, Hang Xu, and Li~Zhang.
\newblock From flatland to space: Teaching vision-language models to perceive and reason in 3d.
\newblock \emph{arXiv preprint arXiv:2503.22976}, 2025{\natexlab{a}}.

\bibitem[Feng et~al.(2025)Feng, Zhang, Li, et~al.]{feng2025onethinker}
Kaituo Feng, Manyuan Zhang, Hongyu Li, et~al.
\newblock Onethinker: All-in-one reasoning model for image and video.
\newblock \emph{arXiv preprint arXiv:2512.03043}, 2025.

\bibitem[Shao et~al.(2024)Shao, Wang, Zhu, et~al.]{deepseekmath2024}
Zhihong Shao, Peiyi Wang, Qihao Zhu, et~al.
\newblock Deepseekmath: Pushing the limits of mathematical reasoning in open language models.
\newblock \emph{arXiv preprint arXiv:2402.03300}, 2024.

\bibitem[Guo et~al.(2025)Guo, Yang, Zhang, et~al.]{guo2025deepseekr1}
Daya Guo, Dejian Yang, Haowei Zhang, et~al.
\newblock Deepseek-r1: Incentivizing reasoning capability in llms via reinforcement learning.
\newblock \emph{arXiv preprint arXiv:2501.12948}, 2025.

\bibitem[Lu et~al.(2026)Lu, Ma, Chen, et~al.]{lu2026visualprimitives}
Ruijie Lu, Yiyang Ma, Xiaokang Chen, et~al.
\newblock Thinking with visual primitives.
\newblock \emph{Technical report}, 2026.

\bibitem[Yuan et~al.(2025{\natexlab{b}})Yuan, Cui, Huang, et~al.]{embodiedr12025}
Yifu Yuan, Haiqin Cui, Yaoting Huang, et~al.
\newblock Embodied-r1: Reinforced embodied reasoning for general robotic manipulation.
\newblock \emph{arXiv preprint arXiv:2508.13998}, 2025{\natexlab{b}}.

\bibitem[Eiter et~al.(1994)Eiter, Mannila, et~al.]{eiter1994computing}
Thomas Eiter, Heikki Mannila, et~al.
\newblock Computing discrete fr{\'e}chet distance.
\newblock 1994.

\bibitem[Ilharco et~al.(2023)Ilharco, Ribeiro, Wortsman, et~al.]{ilharco2023taskarithmetic}
Gabriel Ilharco, Marco~Tulio Ribeiro, Mitchell Wortsman, et~al.
\newblock Editing models with task arithmetic.
\newblock \emph{International Conference on Learning Representations (ICLR)}, 2023.

\bibitem[Agarwal et~al.(2023)Agarwal, Vieillard, Zhou, et~al.]{gkd2023}
Rishabh Agarwal, Nino Vieillard, Yongchao Zhou, et~al.
\newblock On-policy distillation of language models: Learning from self-generated mistakes.
\newblock \emph{arXiv preprint arXiv:2306.13649}, 2023.

\bibitem[Tong et~al.(2024)Tong, Brown~II, Wu, Woo, Iyer, Akula, Yang, Yang, Middepogu, Wang, et~al.]{tong2024cambrian}
Shengbang Tong, Ellis~L Brown~II, Penghao Wu, Sanghyun Woo, Adithya~Jairam Iyer, Sai~Charitha Akula, Shusheng Yang, Jihan Yang, Manoj Middepogu, Ziteng Wang, et~al.
\newblock Cambrian-1: A fully open, vision-centric exploration of multimodal llms.
\newblock In \emph{The Thirty-eighth Annual Conference on Neural Information Processing Systems}, 2024.

\bibitem[Jia et~al.(2025)Jia, Qi, Zhang, Zhang, Yu, He, Wang, and Yi]{jia2025omnispatial}
Mengdi Jia, Zekun Qi, Shaochen Zhang, Wenyao Zhang, Xinqiang Yu, Jiawei He, He~Wang, and Li~Yi.
\newblock Omnispatial: Towards comprehensive spatial reasoning benchmark for vision language models.
\newblock \emph{arXiv preprint arXiv:2506.03135}, 2025.

\bibitem[Song et~al.(2024)Song, Blukis, Tremblay, Tyree, Su, and Birchfield]{song2024robospatial}
Chan~Hee Song, Valts Blukis, Jonathan Tremblay, Stephen Tyree, Yu~Su, and Stan Birchfield.
\newblock Robospatial: Teaching spatial understanding to 2d and 3d vision-language models for robotics.
\newblock \emph{arXiv preprint arXiv:2411.16537}, 2024.

\bibitem[Yeh et~al.(2025)Yeh, Wang, Tong, Cheng, Wang, Chu, Zhai, Chen, Gao, and Ma]{yeh2025allangles}
Chun-Hsiao Yeh, Chenyu Wang, Shengbang Tong, Ta-Ying Cheng, Rouyu Wang, Tianzhe Chu, Yuexiang Zhai, Yubei Chen, Shenghua Gao, and Yi~Ma.
\newblock Seeing from another perspective: Evaluating multi-view understanding in mllms.
\newblock \emph{arXiv preprint arXiv:2504.15280}, 2025.

\bibitem[Majumdar et~al.(2024)Majumdar, Ajay, Zhang, Putta, Yenamandra, Henaff, Silwal, Mcvay, Maksymets, Arnaud, Yadav, Li, Newman, Sharma, Berges, Zhang, Agrawal, Bisk, Batra, Kalakrishnan, Meier, Paxton, Sax, and Rajeswaran]{majumdar2024openeqa}
Arjun Majumdar, Anurag Ajay, Xiaohan Zhang, Pranav Putta, Sriram Yenamandra, Mikael Henaff, Sneha Silwal, Paul Mcvay, Oleksandr Maksymets, Sergio Arnaud, Karmesh Yadav, Qiyang Li, Ben Newman, Mohit Sharma, Vincent Berges, Shiqi Zhang, Pulkit Agrawal, Yonatan Bisk, Dhruv Batra, Mrinal Kalakrishnan, Franziska Meier, Chris Paxton, Sasha Sax, and Aravind Rajeswaran.
\newblock Openeqa: Embodied question answering in the era of foundation models.
\newblock In \emph{Proceedings of the IEEE/CVF Conference on Computer Vision and Pattern Recognition (CVPR)}, 2024.

\bibitem[Plizzari et~al.(2025)Plizzari, Tonioni, Xian, Kulshrestha, and Tombari]{plizzari2025omnia}
Chiara Plizzari, Alessio Tonioni, Yongqin Xian, Achin Kulshrestha, and Federico Tombari.
\newblock Omnia de egotempo: Benchmarking temporal understanding of multi-modal llms in egocentric videos.
\newblock In \emph{Proceedings of the Computer Vision and Pattern Recognition Conference}, pages 24129--24138, 2025.

\bibitem[Fu et~al.(2025)Fu, Dai, Luo, Li, Ren, Zhang, Wang, Zhou, Shen, Zhang, et~al.]{fu2025video}
Chaoyou Fu, Yuhan Dai, Yongdong Luo, Lei Li, Shuhuai Ren, Renrui Zhang, Zihan Wang, Chenyu Zhou, Yunhang Shen, Mengdan Zhang, et~al.
\newblock Video-mme: The first-ever comprehensive evaluation benchmark of multi-modal llms in video analysis.
\newblock In \emph{Proceedings of the IEEE/CVF conference on computer vision and pattern recognition}, pages 24108--24118, 2025.

\bibitem[Li et~al.(2024)Li, Wang, He, et~al.]{li2024mvbench}
Kunchang Li, Yali Wang, Yinan He, et~al.
\newblock Mvbench: A comprehensive multi-modal video understanding benchmark.
\newblock \emph{arXiv preprint arXiv:2311.17005}, 2024.

\bibitem[Wu et~al.(2024{\natexlab{b}})Wu, Li, Chen, et~al.]{wu2024longvideobench}
Haoning Wu, Dongxu Li, Bei Chen, et~al.
\newblock Longvideobench: A benchmark for long-context interleaved video-language understanding.
\newblock \emph{arXiv preprint arXiv:2407.15754}, 2024{\natexlab{b}}.

\bibitem[Zhang et~al.(2025{\natexlab{b}})Zhang, Wang, Ge, Ge, Li, Shan, and Wang]{zhang2025timelens}
Jun Zhang, Teng Wang, Yuying Ge, Yixiao Ge, Xinhao Li, Ying Shan, and Limin Wang.
\newblock Timelens: Rethinking video temporal grounding with multimodal llms.
\newblock \emph{arXiv preprint arXiv:2512.14698}, 2025{\natexlab{b}}.

\bibitem[Lu et~al.(2023)Lu, Fan, Deng, Liu, Li, and Wang]{lu2023vlgrasp}
Yuhao Lu, Yixuan Fan, Beixing Deng, Fangfu Liu, Yali Li, and Shengjin Wang.
\newblock {VL-Grasp}: a {6-Dof} interactive grasp policy for language-oriented objects in cluttered indoor scenes.
\newblock \emph{arXiv preprint arXiv:2308.00640}, 2023.
\newblock URL \url{https://arxiv.org/abs/2308.00640}.
\newblock IROS 2023.

\bibitem[Abdolmaleki et~al.(2025)Abdolmaleki, Abeyruwan, Ainslie, et~al.]{geminirobotics15}
Abbas Abdolmaleki, Saminda Abeyruwan, Joshua Ainslie, et~al.
\newblock Gemini robotics 1.5: Pushing the frontier of generalist robots with advanced embodied reasoning, thinking, and motion transfer.
\newblock \emph{arXiv preprint arXiv:2510.03342}, 2025.

\bibitem[Xue et~al.(2025)Xue, Ge, Zeng, Li, Liu, Chen, and Fan]{xue2025pio}
Haotian Xue, Yunhao Ge, Yu~Zeng, Zhaoshuo Li, Ming-Yu Liu, Yongxin Chen, and Jiaojiao Fan.
\newblock Point-it-out: Benchmarking embodied reasoning for vision language models in multi-stage visual grounding.
\newblock \emph{arXiv preprint arXiv:2509.25794}, 2025.

\bibitem[Windecker et~al.(2025)Windecker, Patel, Reuss, Schwarzkopf, Cadena, Lioutikov, Hutter, and Frey]{windecker2025navitrace}
Tim Windecker, Manthan Patel, Moritz Reuss, Richard Schwarzkopf, Cesar Cadena, Rudolf Lioutikov, Marco Hutter, and Jonas Frey.
\newblock Navitrace: Evaluating embodied navigation of vision-language models.
\newblock \emph{arXiv preprint arXiv:2510.26909}, 2025.

\bibitem[Yue et~al.(2024)Yue, Ni, Zhang, et~al.]{yue2024mmmu}
Xiang Yue, Yuansheng Ni, Kai Zhang, et~al.
\newblock Mmmu: A massive multi-discipline multimodal understanding and reasoning benchmark for expert agi.
\newblock \emph{Proceedings of the IEEE/CVF Conference on Computer Vision and Pattern Recognition (CVPR)}, 2024.

\bibitem[Yu et~al.(2023)Yu, Yang, Li, et~al.]{yu2023mmvet}
Weihao Yu, Zhengyuan Yang, Linjie Li, et~al.
\newblock Mm-vet: Evaluating large multimodal models for integrated capabilities.
\newblock \emph{arXiv preprint arXiv:2308.02490}, 2023.

\bibitem[{xAI}(2024)]{xai2024realworldqa}
{xAI}.
\newblock Grok-1.5 vision preview and realworldqa.
\newblock \url{https://x.ai/news/grok-1.5v}, 2024.
\newblock Accessed 2026-07-30.

\bibitem[Liu et~al.(2023{\natexlab{b}})Liu, Duan, Zhang, et~al.]{liu2023mmbench}
Yuan Liu, Haodong Duan, Yuanhan Zhang, et~al.
\newblock Mmbench: Is your multi-modal model an all-around player?
\newblock \emph{arXiv preprint arXiv:2307.06281}, 2023{\natexlab{b}}.

\bibitem[Zhou et~al.(2023)Zhou, Lu, Mishra, Brahma, Basu, Luan, Zhou, and Hou]{zhou2023ifeval}
Jeffrey Zhou, Tianjian Lu, Swaroop Mishra, Siddhartha Brahma, Sujoy Basu, Yi~Luan, Denny Zhou, and Le~Hou.
\newblock Instruction-following evaluation for large language models.
\newblock \emph{arXiv preprint arXiv:2311.07911}, 2023.

\bibitem[Wang et~al.(2024{\natexlab{b}})Wang, Ma, Zhang, Ni, Chandra, Guo, Ren, Arulraj, He, Jiang, Li, Ku, Wang, Zhuang, Fan, Yue, and Chen]{wang2024mmlupro}
Yubo Wang, Xueguang Ma, Ge~Zhang, Yuansheng Ni, Abhranil Chandra, Shiguang Guo, Weiming Ren, Aaran Arulraj, Xuan He, Ziyan Jiang, Tianle Li, Max Ku, Kai Wang, Alex Zhuang, Rongqi Fan, Xiang Yue, and Wenhu Chen.
\newblock {MMLU-Pro}: A more robust and challenging multi-task language understanding benchmark.
\newblock In \emph{Advances in Neural Information Processing Systems}, 2024{\natexlab{b}}.

\bibitem[Patil et~al.(2025)Patil, Mao, Ji, Yan, Suresh, Stoica, and Gonzalez]{patil2025bfcl}
Shishir~G. Patil, Huanzhi Mao, Charlie Cheng-Jie Ji, Fanjia Yan, Vishnu Suresh, Ion Stoica, and Joseph~E. Gonzalez.
\newblock The berkeley function calling leaderboard (bfcl): From tool use to agentic evaluation of large language models.
\newblock In \emph{Proceedings of the 42nd International Conference on Machine Learning (ICML)}, 2025.

\bibitem[Jain et~al.(2024)Jain, Han, Gu, Li, Yan, Zhang, Wang, Solar-Lezama, Sen, and Stoica]{jain2024livecodebench}
Naman Jain, King Han, Alex Gu, Wen-Ding Li, Fanjia Yan, Tianjun Zhang, Sida Wang, Armando Solar-Lezama, Koushik Sen, and Ion Stoica.
\newblock {LiveCodeBench}: Holistic and contamination-free evaluation of large language models for code.
\newblock \emph{arXiv preprint arXiv:2403.07974}, 2024.

\bibitem[Li et~al.(2026)Li, Sun, Zhang, Kang, Wang, Qiu, Jiang, Cui, and Chen]{deepinsight2026}
Siyi Li, Chunyu Sun, Jiahao Zhang, Yuchen Kang, Wuliang Wang, Yu~Qiu, Rui Jiang, Haitao Cui, and Jie Chen.
\newblock Deepinsight: A unified evaluation infrastructure across the physical ai stack.
\newblock \emph{arXiv preprint arXiv:2606.17574}, 2026.

\bibitem[Bai et~al.(2025{\natexlab{b}})Bai, Cai, Chen, Chen, Chen, Cheng, Deng, Ding, Gao, Ge, et~al.]{bai2025qwen3}
Shuai Bai, Yuxuan Cai, Ruizhe Chen, Keqin Chen, Xionghui Chen, Zesen Cheng, Lianghao Deng, Wei Ding, Chang Gao, Chunjiang Ge, et~al.
\newblock Qwen3-vl technical report.
\newblock \emph{arXiv preprint arXiv:2511.21631}, 2025{\natexlab{b}}.

\bibitem[Kolve et~al.(2017)Kolve, Mottaghi, Han, et~al.]{kolve2017ai2thor}
Eric Kolve, Roozbeh Mottaghi, Winson Han, et~al.
\newblock Ai2-thor: An interactive 3d environment for visual ai.
\newblock \emph{arXiv preprint arXiv:1712.05474}, 2017.

\bibitem[Deitke et~al.(2022)Deitke, VanderBilt, Herrasti, et~al.]{deitke2022procthor}
Matt Deitke, Eli VanderBilt, Alvaro Herrasti, et~al.
\newblock Procthor: Large-scale embodied ai using procedural generation.
\newblock \emph{Advances in Neural Information Processing Systems (NeurIPS)}, 2022.

\bibitem[Sheng et~al.(2024)Sheng, Zhang, Ye, Wu, Zhang, Zhang, Peng, Lin, and Wu]{sheng2024hybridflow}
Guangming Sheng, Chi Zhang, Zilingfeng Ye, Xibin Wu, Wang Zhang, Ru~Zhang, Yanghua Peng, Haibin Lin, and Chuan Wu.
\newblock Hybridflow: A flexible and efficient rlhf framework.
\newblock \emph{arXiv preprint arXiv: 2409.19256}, 2024.

\bibitem[Shoeybi et~al.(2019)Shoeybi, Patwary, Puri, LeGresley, Casper, and Catanzaro]{megatron-lm}
Mohammad Shoeybi, Mostofa Patwary, Raul Puri, Patrick LeGresley, Jared Casper, and Bryan Catanzaro.
\newblock Megatron-lm: Training multi-billion parameter language models using model parallelism.
\newblock \emph{arXiv preprint arXiv:1909.08053}, 2019.

\bibitem[Kwon et~al.(2023)Kwon, Li, Zhuang, Sheng, Zheng, Yu, Gonzalez, Zhang, and Stoica]{kwon2023efficient}
Woosuk Kwon, Zhuohan Li, Siyuan Zhuang, Ying Sheng, Lianmin Zheng, Cody~Hao Yu, Joseph~E. Gonzalez, Hao Zhang, and Ion Stoica.
\newblock Efficient memory management for large language model serving with pagedattention.
\newblock In \emph{Proceedings of the ACM SIGOPS 29th Symposium on Operating Systems Principles}, 2023.

\end{thebibliography}

\appendix
\FloatBarrier
\newpage
\section{Training and Evaluation Details}
\label{app:training_details}
\footnotesize

Table~\ref{tab:appendix_training} reports the available frozen 35B-A3B GRPO manifests. The 2B track uses the same verifier and optimization interfaces with branch-level settings recorded in its own run manifests. Training uses the verl~\citep{sheng2024hybridflow} and Megatron-LM~\citep{megatron-lm} stack with vLLM~\citep{kwon2023efficient} rollouts and bfloat16 precision.

\begin{table}[H]
\centering
\scriptsize
\setlength{\tabcolsep}{3.5pt}
\renewcommand{\arraystretch}{1.08}
\caption{Recorded 35B-A3B specialist GRPO settings. Affordance and trajectory are two optimization contracts within the Guidance expert.}
\label{tab:appendix_training}
\begin{tabularx}{0.98\linewidth}{>{\raggedright\arraybackslash}Xrrrr}
\toprule
\textbf{Setting} & \textbf{Spatial} & \shortstack{\textbf{Guidance}\\\textbf{(affordance)}} & \shortstack{\textbf{Temporal}} & \shortstack{\textbf{Guidance}\\\textbf{(trajectory)}} \\
\midrule
Learning rate & $1{\times}10^{-6}$ & $1{\times}10^{-6}$ & $1{\times}10^{-6}$ & $1{\times}10^{-6}$ \\
Rollouts per prompt & 8 & 8 & 8 & 8 \\
Actor KL coefficient & 0.01 & 0.01 & 0.01 & off \\
Global batch size & 128 & 128 & 64 & 64 \\
Maximum prompt length & 8{,}192 & 5{,}120 & 10{,}240 & 10{,}240 \\
Maximum response length & 5{,}120 & 5{,}120 & 5{,}120 & 8{,}192 \\
Train parallelism & TP2/EP8 & TP2/EP8 & TP2/EP8 & TP2/EP8 \\
Rollout tensor parallelism & 8 & 4 & 4 & 4 \\
\bottomrule
\end{tabularx}
\end{table}

The policy-space integration run initializes the student from the TIES-merged checkpoint and uses a global batch size of 240, learning rate $1{\times}10^{-6}$, two student rollouts per prompt, maximum prompt and response lengths of 10{,}240 and 5{,}120, and one training epoch. It uses 12 actor GPUs with TP2/EP4 and rollout TP2. The same four expert checkpoints used by TIES are served as frozen MOPD teachers. Each training contract is routed to its responsible expert; the run manifest fixes the contract-to-expert mapping, teacher identifiers, and route proportions.

\subsection{Evaluation Prompts}
\label{app:deepinsight_protocol}

All public benchmarks are run with DeepInsight~\citep{deepinsight2026}. For each benchmark, the media input, question, prompt, parser, and scorer are fixed across models. \xvista and the corresponding Qwen checkpoint use a 128K context window, at most 64 video frames, temperature 0.7, top-$p$ 0.95, top-$k$ 20, and at most 16{,}384 output tokens. Other model families use the inference settings recommended by their papers or model cards. Following the benchmark-by-benchmark presentation of RynnBrain~\citep{rynnbrain2026}, Table~\ref{tab:evaluation_prompts} lists the text templates used in our evaluation. Media is supplied before the text and is omitted from the templates below.

\begingroup
\footnotesize
\setlength{\tabcolsep}{4pt}
\renewcommand{\arraystretch}{1.08}
\begin{longtable}{@{}>{\raggedright\arraybackslash}p{0.20\linewidth}>{\raggedright\arraybackslash\ttfamily}p{0.74\linewidth}@{}}
\caption{Prompt templates used for public benchmark evaluation. Braced terms are fields supplied by the dataset.}
\label{tab:evaluation_prompts}\\
\toprule
\textbf{Benchmark} & \normalfont\textbf{Prompt template} \\
\midrule
\endfirsthead
\multicolumn{2}{r}{\scriptsize Continued from the previous page} \\
\toprule
\textbf{Benchmark} & \normalfont\textbf{Prompt template} \\
\midrule
\endhead
\bottomrule
\endfoot

General
& \{question\} \\
\midrule

LongVideoBench
& \{question\}\newline
  \{options\}\newline
  Answer with the option letter from the given choices directly. \\
\midrule



RoboRefIt
& Output the target location with the form of a bounding box.\newline
  Output format: [\{"bbox\_2d": [x1, y1, x2, y2], "label": "label"\}].\newline
  If no target is found, respond with an empty list.\newline
  Question: \{question\} \\
\midrule

ShareRobot-Affordance
& \{question\}\newline
  Return only the affordance bounding box:\newline
  [\{"bbox\_2d": [x1, y1, x2, y2], "label": "affordance area"\}] \\
\midrule

PixMoPointsEval
& Locate all instances of \{question\} in the image.\newline
  Output point coordinates as a JSON list using 0--1000 coordinates.\newline
  If no instance is found, respond Not found. \\
\midrule

PIO
& Given the image and language description, output one to three points that best localize the description.\newline
  Output format: [\{"point\_2d": [x, y], "label": "point\_1"\}].\newline
  Language description: \{question\} \\
\midrule

VABench-point
& \{question\}\newline
  Answer with a list of 2D points in JSON format.\newline
  Output format: [\{"point\_2d": [x, y], "label": "label"\}] \\
\midrule

where2place
& \{question\}\newline
  Return several valid points inside the requested vacant area as a JSON array.\newline
  Coordinates are normalized between 0 and 1000.\newline
  Return only the JSON array. \\
\midrule

PointBench
& \{hint\}\newline
  Return the requested point coordinates as a JSON array.\newline
  Coordinates are normalized between 0 and 1000.\newline
  For counting questions, return one point for each target instance.\newline
  Return only the JSON array. \\
\midrule

VABench-trace / ShareRobot-Trajectory
& \{question\}\newline
  Answer with a list of 2D points in JSON format.\newline
  Output format: [\{"point\_2d": [x, y], "label": "waypoint N"\}, ...] \\
\midrule

NaviTrace
& System: Act as a navigation expert. Start near the bottom center, adapt the path to the specified embodiment, stop for unsafe traffic conditions, and output only 2D points in JSON format.\newline
  Embodiment: \{hint\}\newline
  Task: \{question\}\newline
  Predict a feasible path forward from the current first-person view. \\
\midrule

RoboSpatial-Home
& \{question\}\newline
  For pinpoint or localization questions, return normalized points as a JSON array.\newline
  For yes/no questions, answer only Yes or No. \\
\midrule

RoboSpatial judge
& Question: \{question\}\newline
  Reference answer: \{gold\}\newline
  Model prediction: \{pred\}\newline
  Return only Correct or Incorrect. \\

\end{longtable}
\endgroup

Before scoring, we remove the \texttt{<think>} span and unwrap an optional \texttt{<answer>} block. Bounding-box grounding rows use IoU@0.5 in $[0,1000]$ coordinates, while ShareRobot-Affordance reports mean IoU over predicted and reference affordance boxes with invalid or missing predictions assigned zero. Pointing rows retain their benchmark-native scorers: PixMoPointsEval uses assignment-based point-level F1, and PIO uses point-in-mask scoring. Trajectory rows are not collapsed to a single transform: NaviTrace keeps its official semantic-aware normalized protocol~\citep{windecker2025navitrace}, VABench-trace reports trajectory RMSE, and ShareRobot-Trajectory reports raw discrete Fr\'echet distance. 


\subsection{Capek-StateBench Protocol}
\label{app:statebench_protocol}

Capek-StateBench contains two 500-example tracks aligned with the State Verification training objectives. Capek-StateBench-P evaluates physical-state verification from local image or video evidence. Predictions are normalized to the state label licensed by the question and scored by exact match; invalid or ambiguous answers receive zero. The primary PSV score is the mean accuracy over the complete split.

Capek-StateBench-T evaluates progress-value estimation from a high-level task instruction and an image-based causal visual prefix ending at the current checkpoint; future observations are excluded. The split contains 213 task-condition records and 287 primitive-skill records. For task-condition records, progress is the percentage of satisfied BEHAVIOR-1K goal conditions, while for primitive-skill records it is the percentage of completed annotated child skills, including navigation:
\begin{equation}
p_t^{\mathrm{cond}}
=
100\frac{|\mathcal{G}_t^{\mathrm{sat}}|}{|\mathcal{G}|},
\qquad
p_t^{\mathrm{skill}}
=
100\frac{n_t}{N}.
\end{equation}
The two definitions remain separately typed. The next-action target is the immediate annotated continuation of the reference execution, rather than a claim of an optimal policy action.

For a valid progress prediction $\hat{p}_i\in[0,100]$, the progress score is
\begin{equation}
s_i^{\mathrm{prog}}
=
\max\left(0,\,1-\frac{|\hat{p}_i-p_i|}{25}\right);
\end{equation}
missing, malformed, or out-of-range values receive zero. Next actions are scored by a fixed semantic judge that accepts harmless paraphrases but requires the operation, object, target, direction, and relation to agree with the reference. Letting $s_i^{\mathrm{act}}\in\{0,1\}$ denote this result, the task-track score is
\begin{equation}
s_i^{\mathrm{task}}
=
\frac{5s_i^{\mathrm{prog}}+4s_i^{\mathrm{act}}}{9}.
\end{equation}
The primary Capek-StateBench-T result is the mean over all 500 records; results by progress annotation type, scenario type, and terminal status are retained as diagnostics. Full prompts and parser implementations are provided in the released DeepInsight configurations.

\subsection{Trajectory Reward and VABench-Trace Score}
\label{app:trajectory_scoring}

For nonempty predicted waypoints $P=(p_1,\ldots,p_n)$ and reference waypoints $Q=(q_1,\ldots,q_m)$, coordinates are first mapped to $[0,1]^2$. Both paths are then resampled at uniform arc-length positions to a common length, producing $\bar P$ and $\bar Q$; a zero-length constant path is represented by repeating its single location. A coupling $\gamma=((i_\ell,j_\ell))_{\ell=1}^{L}$ starts at $(1,1)$, ends at the final index pair, and has increments in $\{(1,0),(0,1),(1,1)\}$. Let $\Gamma$ be the set of these endpoint-preserving monotone couplings. The discrete Fr\'echet distance is
\begin{equation}
D_{\mathrm{DFD}}(\bar P,\bar Q)=
\min_{\gamma\in\Gamma}
\max_{(i,j)\in\gamma}\lVert \bar p_i-\bar q_j\rVert_2.
\label{eq:dfd}
\end{equation}
Equation~\ref{eq:trajectory_reward} applies the frozen geometric kernel to this distance. Invalid or empty parsed trajectories receive zero. The constant-path convention above makes resampling total on every nonempty valid path. Arc-length resampling removes a spurious dependence on waypoint count while the discrete Fr\'echet distance preserves path order and worst-case alignment.

VABench-trace uses a separate evaluation path. Its recorded evaluator applies the benchmark coordinate normalization and path preprocessing to each episode, maps trajectories into the padded-square pixel coordinate system, and resamples a predicted path when its waypoint count differs from the reference. The public table reports the global trajectory RMSE over all valid waypoint pairs,
\begin{equation}
S_{\mathrm{VABench}\text{-}\mathrm{trace}}
=
\sqrt{\frac{1}{M}\sum_{j=1}^{M}\lVert \hat q_j-q_j\rVert_2^2},
\label{eq:vabench_trace_score}
\end{equation}
where $M$ is the total number of compared waypoints after preprocessing. This value is lower-is-better and remains in the benchmark pixel coordinate scale. The training reward remains the resampled-path kernel in Equation~\ref{eq:trajectory_reward}; it is not used to transform the public VABench-trace table entries.

\normalsize
\clearpage
\section{Qualitative EmbodiedBench Rollout}
\label{app:embodiedbench_rollouts}

\begin{table}[H]
\centering
\small
\setlength{\tabcolsep}{4pt}
\renewcommand{\arraystretch}{1.08}
\caption{Selected EmbodiedBench replay sample.}
\label{tab:embodiedbench_alf_ep10}
\begin{tabularx}{\linewidth}{@{}>{\bfseries}lX>{\bfseries}lX@{}}
\toprule
Sample & \texttt{EB-ALF / long\_horizon / episode-10} &
Family & \texttt{long-horizon manipulation} \\
Instruction & Pick up knife, slice apple, put knife in bowl, heat slice of apple in microwave, put apple slice on table. &
Outcome & 20 steps / 0 invalid / success \\
\bottomrule
\end{tabularx}
\end{table}

Figure~\ref{fig:embodiedbench_alf_ep10_grid} shows a successful long-horizon EmbodiedBench episode. The model first resolves the required tool and object, slices the apple, stores the knife in a bowl, loads and heats the apple slice in the microwave, and finally places it on the dining table. The episode illustrates how the same execution-centric interface combines object grounding, state-changing manipulation, appliance interaction, and final task-state verification within one rollout.

\begingroup
\newcommand{\embodiedframe}[3]{%
\begin{minipage}[t]{0.188\linewidth}
\centering
\includegraphics[width=\linewidth]{#1}\\[-0.15em]
{\scriptsize\textbf{#2}\par}
{\scriptsize\ttfamily\begin{tabular}{@{}c@{}}#3\end{tabular}\par}
\end{minipage}}

\begin{figure}[H]
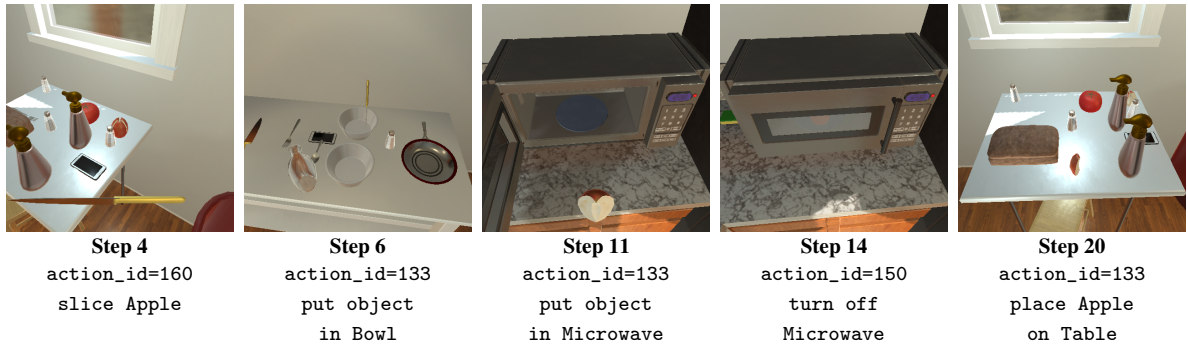

\centering
\setlength{\tabcolsep}{2pt}
\renewcommand{\arraystretch}{1.18}
\begin{tabular}{@{}ccccc@{}}
\embodiedframe{embodied_showcase/alf_ep10_step_04_slice_apple.png}{Step 4}{action\_id=160\\slice Apple} &
\embodiedframe{embodied_showcase/alf_ep10_step_06_knife_in_bowl.png}{Step 6}{action\_id=133\\put object\\in Bowl} &
\embodiedframe{embodied_showcase/alf_ep10_step_11_load_microwave.png}{Step 11}{action\_id=133\\put object\\in Microwave} &
\embodiedframe{embodied_showcase/alf_ep10_step_14_heat.png}{Step 14}{action\_id=150\\turn off\\Microwave} &
\embodiedframe{embodied_showcase/alf_ep10_step_20_deliver_to_table.png}{Step 20}{action\_id=133\\place Apple\\on Table} \\
\end{tabular}
\caption{\textbf{Task-critical EmbodiedBench keyframes.} Each selected frame is paired with the model action executed at that step. The full action trace contains 20 successful steps with no invalid action.}
\label{fig:embodiedbench_alf_ep10_grid}
\end{figure}
\endgroup

\begin{table}[H]
\centering
\scriptsize
\setlength{\tabcolsep}{4pt}
\renewcommand{\arraystretch}{1.05}
\caption{Execution outcome summary for Figure~\ref{fig:embodiedbench_alf_ep10_grid}.}
\label{tab:embodiedbench_alf_ep10_feedback}
\begin{tabularx}{\linewidth}{@{}p{0.12\linewidth}p{0.42\linewidth}X@{}}
\toprule
\textbf{Step(s)} & \textbf{Model action sequence} & \textbf{Outcome summary} \\
\midrule
1--4 & Find knife, pick it up, find apple, slice apple. & All calls execute successfully; the apple is transformed into a sliced state while the knife remains in hand. \\
5--6 & Find bowl, put down the held object. & The knife is placed in the bowl, clearing the hand for the apple slice. \\
7--14 & Find and pick up apple, find microwave, open it, place apple inside, close it, turn it on, turn it off. & The apple slice is loaded into the microwave and the appliance interaction completes without invalid actions. \\
15--20 & Open microwave, find and pick up apple, close microwave, find dining table, put down the held object. & The heated apple slice is retrieved and placed on the dining table; the final environment state satisfies the task. \\
\bottomrule
\end{tabularx}
\end{table}

\begin{table}[H]
\centering
\scriptsize
\setlength{\tabcolsep}{3pt}
\renewcommand{\arraystretch}{1.04}
\caption{Complete model action trace for the EmbodiedBench episode. Every step returns \texttt{Last action executed successfully.}}
\label{tab:embodiedbench_alf_ep10_trace}
\begin{tabularx}{\linewidth}{@{}>{\bfseries}rX>{\bfseries}rX@{}}
\toprule
\textbf{Step} & \textbf{Action} & \textbf{Step} & \textbf{Action} \\
\midrule
1 & \texttt{action\_id=34}: find Knife & 11 & \texttt{action\_id=133}: put apple in Microwave \\
2 & \texttt{action\_id=127}: pick up Knife & 12 & \texttt{action\_id=144}: close Microwave \\
3 & \texttt{action\_id=22}: find Apple & 13 & \texttt{action\_id=149}: turn on Microwave \\
4 & \texttt{action\_id=160}: slice Apple & 14 & \texttt{action\_id=150}: turn off Microwave \\
5 & \texttt{action\_id=16}: find Bowl & 15 & \texttt{action\_id=143}: open Microwave \\
6 & \texttt{action\_id=133}: put knife in Bowl & 16 & \texttt{action\_id=22}: find Apple \\
7 & \texttt{action\_id=22}: find Apple & 17 & \texttt{action\_id=107}: pick up Apple \\
8 & \texttt{action\_id=107}: pick up Apple & 18 & \texttt{action\_id=144}: close Microwave \\
9 & \texttt{action\_id=38}: find Microwave & 19 & \texttt{action\_id=18}: find DiningTable \\
10 & \texttt{action\_id=143}: open Microwave & 20 & \texttt{action\_id=133}: place apple on DiningTable \\
\bottomrule
\end{tabularx}
\end{table}

\FloatBarrier
\clearpage
\section{Qualitative VIGIL Rollouts}
\label{app:vigil_rollouts}

\begin{table}[H]
\centering
\small
\setlength{\tabcolsep}{4pt}
\renewcommand{\arraystretch}{1.08}
\caption{Selected complex \vigil replay sample.}
\label{tab:vigil_complex_sample}
\begin{tabularx}{\linewidth}{@{}>{\bfseries}lX>{\bfseries}lX@{}}
\toprule
Sample & \texttt{sample-627} &
Family & \texttt{T2\_search\_then\_interact} \\
Instruction & Find the toaster, turn it on, then report success. &
Score & 0.9543 / success \\
\bottomrule
\end{tabularx}
\end{table}

Figure~\ref{fig:vigil_toaster_grid} shows a representative \vigil episode in which the model first approaches the table, identifies the toaster among nearby objects, and repeatedly attempts to activate it. Several early activation calls do not change the task state; in the rollout, the model attributes these failures to being slightly out of interaction range or to imprecise targeting. This distance diagnosis is inferred by the model from the unchanged visual and task-state evidence, rather than returned as an explicit environment message. The model then moves closer, successfully toggles the toaster at Step~8, and reports success after the activated state becomes visible. This trajectory highlights a multi-step recovery behavior: the model does not stop after the first failed interaction, but continues to refine its position and action until the task state is achieved.

\begingroup
\newcommand{\vigilframe}[3]{%
\begin{minipage}[t]{0.235\linewidth}
\centering
\includegraphics[width=0.8\linewidth]{#1}\\[-0.15em]
{\scriptsize\textbf{#2}\par}
{\scriptsize\ttfamily\begin{tabular}{@{}c@{}}#3\end{tabular}\par}
\end{minipage}}

\begin{figure}[H]
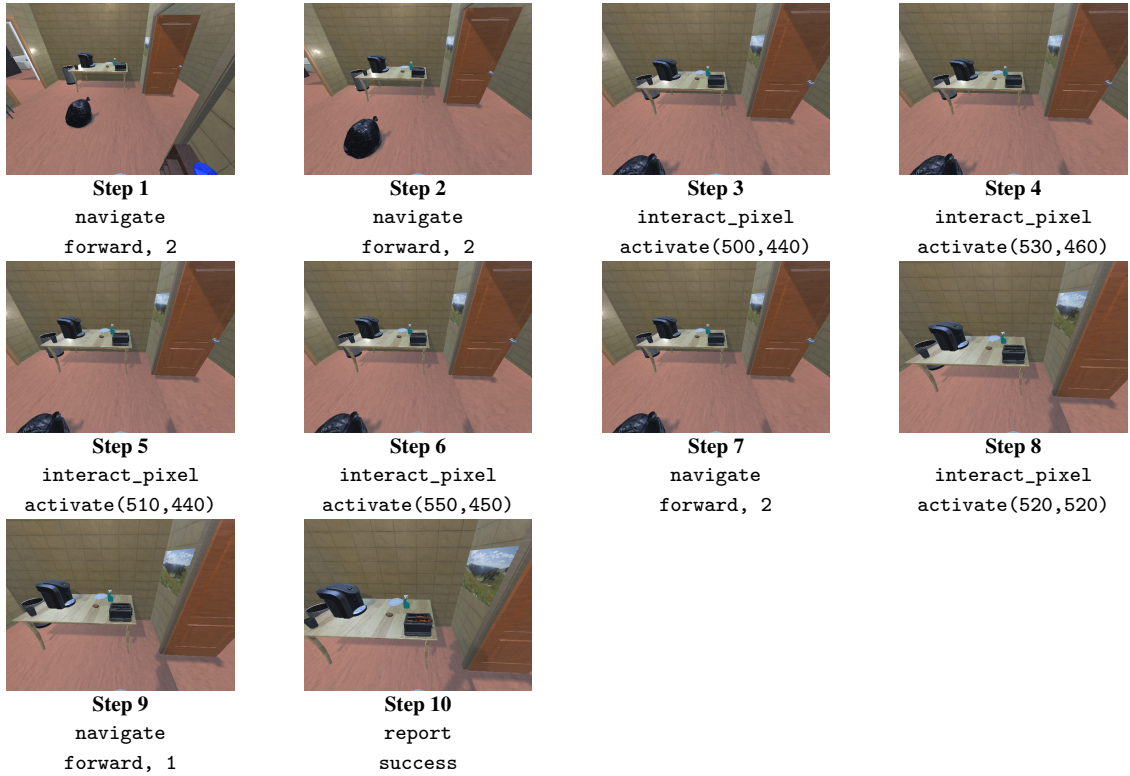

\centering
\setlength{\tabcolsep}{2.5pt}
\renewcommand{\arraystretch}{1.18}
\begin{tabular}{@{}cccc@{}}
\vigilframe{vigil_showcase/toaster_step_01.png}{Step 1}{navigate\\forward, 2} &
\vigilframe{vigil_showcase/toaster_step_02.png}{Step 2}{navigate\\forward, 2} &
\vigilframe{vigil_showcase/toaster_step_03.png}{Step 3}{interact\_pixel\\activate(500,440)} &
\vigilframe{vigil_showcase/toaster_step_04.png}{Step 4}{interact\_pixel\\activate(530,460)} \\
\vigilframe{vigil_showcase/toaster_step_05.png}{Step 5}{interact\_pixel\\activate(510,440)} &
\vigilframe{vigil_showcase/toaster_step_06.png}{Step 6}{interact\_pixel\\activate(550,450)} &
\vigilframe{vigil_showcase/toaster_step_07.png}{Step 7}{navigate\\forward, 2} &
\vigilframe{vigil_showcase/toaster_step_08.png}{Step 8}{interact\_pixel\\activate(520,520)} \\
\vigilframe{vigil_showcase/toaster_step_09.png}{Step 9}{navigate\\forward, 1} &
\vigilframe{vigil_showcase/toaster_step_10.png}{Step 10}{report\\success} &
\multicolumn{2}{c}{} \\
\end{tabular}
\caption{\textbf{Step-by-step VIGIL rollout.} Each frame is paired with the exact model function call for that observation. The successful activation occurs at Step 8; Step 10 closes the episode with a matching success report.}
\label{fig:vigil_toaster_grid}
\end{figure}
\endgroup

\begin{table}[H]
\centering
\scriptsize
\setlength{\tabcolsep}{4pt}
\renewcommand{\arraystretch}{1.05}
\caption{Execution outcome summary for the function calls in Figure~\ref{fig:vigil_toaster_grid}.}
\label{tab:vigil_toaster_feedback}
\begin{tabularx}{\linewidth}{@{}p{0.14\linewidth}X@{}}
\toprule
\textbf{Step(s)} & \textbf{Outcome summary} \\
\midrule
1--2 & The agent moves toward the table; no target interaction has been credited yet. \\
3--5 & The calls are aimed at the toaster, but no activation is credited. The model infers from the unchanged state that it may still be too far away or imprecisely targeted; the environment does not expose this distance explanation as feedback. \\
6 & The call appears to land away from the toaster, so no activation is credited. \\
7 & The agent moves closer after the failed activation attempts. \\
8 & The activation call hits the Toaster and the verifier marks the required object state as reached. \\
9--10 & The state remains reached; the final \texttt{report(success)} matches the world state. \\
\bottomrule
\end{tabularx}
\end{table}

\FloatBarrier
\clearpage
\section{Qualitative Static QA Rollouts}
\label{app:static_qa_rollouts}

Each static QA example is shown as a paired visualization and model rollout. The selected samples cover spatial reasoning, point grounding, manipulation trajectory prediction, trace navigation, temporal video understanding, and general chart reading while keeping the evidence close to the corresponding image.

\begingroup
\newcommand{\staticqarollout}[8]{%
\begin{tcolorbox}[
  enhanced,
  breakable,
  colback=white,
  colframe=black!18,
  boxrule=0.35pt,
  arc=1pt,
  left=5pt,
  right=5pt,
  top=4pt,
  bottom=4pt,
  before skip=5pt,
  after skip=5pt
]
{\footnotesize\textbf{#1}\par\vspace{0.25em}}
\noindent
\begin{minipage}[t]{0.34\linewidth}
\vspace{0pt}
\centering
\includegraphics[width=\linewidth,height=#2,keepaspectratio]{#3}\\[0.15em]
{\scriptsize #4\par}
\vspace{0.3em}
{\tiny\raggedright
\textbf{Question.} #5\par
\textbf{Ground truth.} #6\par}
\end{minipage}\hfill
\begin{minipage}[t]{0.63\linewidth}
\vspace{0pt}
\lstinputlisting[style=rolloutcompact,#8]{#7}
\end{minipage}\par
\end{tcolorbox}
}

\staticqarollout
{Physical-state verification: \texttt{StateBench-P / PSVMIX\_21d9464a}}
{0.13\textheight}
{static_qa_showcase/statebench_psv_lighter.png}
{Across the two chronological observations, the model tracks the lighter's final state and returns the correct state label.}
{Based on observations, is the lighter on or off?}
{\texttt{off}.}
{figures/static_qa_showcase/rollouts/statebench_psv_lighter.txt}
{}

\staticqarollout
{Task-state verification: \texttt{StateBench-T / B1K\_CPV\_8d5e714d}}
{0.16\textheight}
{static_qa_showcase/statebench_pve_packing_box.png}
{The model correctly tracks the multi-object packing state and identifies opening the refrigerator door as the next action, but underestimates completion (75\% vs.\ 83.33\%) because of an inconsistent goal-condition count.}
{Task: Put both apple halves, the club sandwich, and the chocolate chip cookie from the chopping board on the kitchen countertop into the packing box on the countertop. Then take the bottle of tea out of the refrigerator, put it into the same box, and close the refrigerator when you're done. Estimate completed task percentage and predict the immediate next action.}
{\texttt{\{"progress": 83.33, "next\_action": "open the fridge door"\}}.}
{figures/static_qa_showcase/rollouts/statebench_pve_packing_box.txt}
{}

\newpage

\staticqarollout
{Spatial reasoning: \texttt{OpenEQA / 1122}}
{0.16\textheight}
{static_qa_showcase/spatial_openeqa_1122.png}
{The model searches across bathroom views and identifies the shower shelf as a feasible place for shampoo.}
{Where can I place a bottle of shampoo while I shower?}
{On the tray in the shower area or on the bathtub ledge.}
{figures/static_qa_showcase/rollouts/spatial_openeqa_1122.txt}
{}

\staticqarollout
{Grounding and trajectory: \texttt{VABench\_trace / fsd\_pointdroid\_4060}}
{0.21\textheight}
{static_qa_showcase/ground_vabench_trace_yellow_to_pot.png}
{The model identifies the yellow object and silver pot, then emits an eight-point object-centric trajectory.}
{Put the yellow object into the silver pot; output eight ordered object-centric waypoints.}
{\texttt{[(652,500), (619,486), (588,427), (547,366), (494,334), (450,373), (406,388), (396,394)]}}
{figures/static_qa_showcase/rollouts/ground_vabench_trace_4060.txt}
{firstline=40}

\newpage
\staticqarollout
{Pointing: \texttt{PIO / 47}}
{0.16\textheight}
{static_qa_showcase/pointing_pio_gripper_handoff_47.png}
{The model grounds an execution handoff by selecting the grasp point where the receiving gripper should act.}
{The left gripper is passing the device to the right gripper. Where should the right gripper act to grasp the device?}
{Valid grasp region shown in green; model point shown in red; Capek score: \texttt{acc=1.0}.}
{figures/static_qa_showcase/rollouts/pointing_pio_47.txt}
{}

\staticqarollout
{Trace navigation: \texttt{NaviTrace / 9d8a...}}
{0.18\textheight}
{static_qa_showcase/trace_navitrace_right_tunnel_1241.png}
{The model chooses the right tunnel and produces a curved path that stays close to the official reference trajectory.}
{Take the right tunnel.}
{Official best-matching trajectory shown in green; model trajectory shown in red; \texttt{dfd=0.045}.}
{figures/static_qa_showcase/rollouts/trace_navitrace_right_tunnel_1241.txt}
{}

\newpage
\staticqarollout
{Video and temporal: \texttt{EgoTempo / grapes}}
{0.16\textheight}
{static_qa_showcase/video_egotempo_grapes.png}
{The model identifies what happens immediately after the grapes are weighed by following the frame sequence.}
{What happens after the person weighs the grapes?}
{The person puts the grapes into a plastic bag and then places it in the shopping basket.}
{figures/static_qa_showcase/rollouts/video_egotempo_grapes.txt}
{}

\staticqarollout
{General: \texttt{InfoVQA\_VAL / 95160}}
{0.24\textheight}
{static_qa_showcase/general_infovqa_price_spike_95160.png}
{The model locates the 2010--2012 price-spike region and answers that the relevant cause is the Arab Spring.}
{What caused unpredictable price spikes between 2010 and 2012?}
{Arab Spring.}
{figures/static_qa_showcase/rollouts/general_infovqa_95160.txt}
{}

\clearpage
\FloatBarrier
\subsection{Static QA Failure Cases}
\label{app:static_qa_failures}

The examples below show representative failures under the same static-QA evaluation interface. In each case, the rollout remains interpretable: the model identifies plausible visual evidence, but commits to an incorrect execution target or spatial relation. These cases complement the successful samples above by showing where execution-facing supervision still exposes brittle decisions.

\staticqarollout
{Failure: \texttt{PIO / 432}}
{0.16\textheight}
{static_qa_showcase/failure_pio_oven_door_432.png}
{The model correctly identifies the oven but places the action point on the lower door surface rather than the annotated handle.}
{Where should I apply force to open the oven door?}
{GT mask in green; parsed model point in red, \texttt{[267, 638]}; \texttt{acc=0.0000}.}
{figures/static_qa_showcase/rollouts/failure_pio_oven_door_432.txt}
{}

\clearpage
\staticqarollout
{Failure: \texttt{VABench-point / fsd\_pointbridge\_94165}}
{0.18\textheight}
{static_qa_showcase/failure_vabench_rabbit_drawer_94165.png}
{The model finds the toy rabbit but grounds the destination near the drawer front instead of the top surface of the drawer.}
{Move the toy rabbit on top of the drawer.}
{GT mask in green; parsed model point in red, \texttt{[237, 868]}; \texttt{acc=0.0000}.}
{figures/static_qa_showcase/rollouts/failure_vabench_rabbit_drawer_94165.txt}
{}

\staticqarollout
{Failure: \texttt{InfoVQA\_VAL / 93846}}
{0.21\textheight}
{static_qa_showcase/failure_infovqa_avengers_93846.png}
{The model reads the Phase One release label and ignores the infographic's Roman-numeral chronological ordering.}
{In phase 1, which movie comes first in the chronological order?}
{GT: \texttt{Captain America: The First Avenger}; parsed answer: \texttt{Iron Man}.}
{figures/static_qa_showcase/rollouts/failure_infovqa_avengers_93846.txt}
{}
\endgroup

\FloatBarrier




\end{document}